\documentclass[manuscript,screen]{acmart}

\usepackage{arydshln}
\usepackage{diagbox}
\usepackage{algorithmic}
\usepackage{algorithm}
\usepackage{multirow}
\usepackage{multicol}
\usepackage{adjustbox}
\usepackage{booktabs}
\usepackage{cleveref}
\usepackage{subfig}
\usepackage{balance}
\usepackage{graphicx}
\usepackage{enumitem}
\usepackage{xcolor}
\usepackage{xspace}
\usepackage{soul}

\makeatletter
\DeclareRobustCommand\onedot{\futurelet\@let@token\@onedot}
\def\@onedot{\ifx\@let@token.\else.\null\fi\xspace}
\DeclareRobustCommand\nodot{\futurelet\@let@token\@nodot}
\def\@nodot{\ifx\@let@token.\else~\null\fi\xspace}
\def\eg{\emph{e.g}\onedot} 

\def\ie{\emph{i.e}\onedot} 

\def\etal{\emph{et al}\onedot}
\makeatother
\newcommand{\rev}[1]{\textcolor{black}{#1}}
\AtBeginDocument{%
  }

\setcopyright{none}
\renewcommand\footnotetextcopyrightpermission[1]{}

\begin{document}

\newcommand{\name}{\textsf{GSVC}}
\title{{\name}: Efficient Video Representation and Compression Through 2D Gaussian Splatting}

\author{Longan Wang}
\affiliation{%
  \institution{National University of Singapore}
  \city{Singapore}\country{Singapore}}
\email{wanglongan@comp.nus.edu.sg}
\orcid{0009-0001-9989-3739}

\author{Yuang Shi}
\affiliation{%
  \institution{National University of Singapore}
  \city{Singapore}\country{Singapore}}
\email{yuangshi@comp.nus.edu.sg}
\orcid{0000-0002-7893-8512}

\author{Wei Tsang Ooi}
\authornote{Corresponding author}
\affiliation{%
  \institution{National University of Singapore}
  \city{Singapore}\country{Singapore}}
\email{ooiwt@comp.nus.edu.sg}
\orcid{0000-0001-8994-1736}

\renewcommand{\shortauthors}{L. Wang, Y. Shi, and W. T. Ooi}

\begin{abstract}
In this work, we explore 2D Gaussian splat as a new primitive for representing and compressing videos, showing its promise as an alternative to traditional signal-based primitives.
We introduce {\name}, a new Gaussian splatting-based video codec that incorporates common techniques used in classical video codecs to exploit temporal redundancy between nearby frames.
In {\name}, each frame is represented by a set of overlapping Gaussian splats, which, when rendered together, approximate the original frame.
\rev{Gaussian splats for the first frame and subsequent frames following scene changes (called anchor frames) are learned from scratch, while the remaining frames (called propagation frames) are initialized from the Gaussian splats of their previous frames and then fine-tuned.}
Scene changes are detected through loss differences during the training process. 
To capture large motion or newly appeared objects, we randomly add Gaussian splats to propagation frames before fine-tuning. 
{\name} supports a knob for rate-distortion trade-offs through controlling the number of Gaussian splats for each frame.  
Gaussian splats with low contributions to the frame quality can be removed during training. 
{\name} supports \rev{two-pass encoding} by propagating the Gaussian splats from a future propagation frame back to an anchor frame, further improving the video quality for challenging video scenes.
To achieve efficient compression, {\name} uses a residual-based Gaussian splats quantization framework that encodes only changes to the Gaussian parameters between frames. 
\rev{Experimental results show that {\name} achieves rate-distortion performance comparable to state-of-the-art video codec implementations for AV1 and VVC.}
GSVC inherits the advantages of Gaussian splatting in rendering, allowing it to achieve the rendering speed of 1500 fps for a $1920\times 1080$ video. 
Project page: \url{https://longanwang-cs.github.io/gsvc/}.
\end{abstract}



\maketitle

\section{Introduction}
\rev{Signal processing techniques, particularly frequency domain signals, serve as the basis for current digital video coding standards~\cite{DBLP:journals/tcsv/WiegandSBL03, DBLP:journals/tcsv/SullivanOHW12, DBLP:conf/vcip/Sullivan20}.}
The first discrete cosine transform (DCT)-based video codec, H.261, was standardized in 1988.
Despite significant advances in compression efficiency through the efforts of the industry and academia, modern standard codecs such as H.266 and AV1 remain DCT-centric.

\begin{figure}[t]
    \centering
    \subfloat[$N$ = 10 ($\mathrm{PSNR} = 12.3\,\mathrm{dB}$)]{\includegraphics[width=0.24\textwidth]{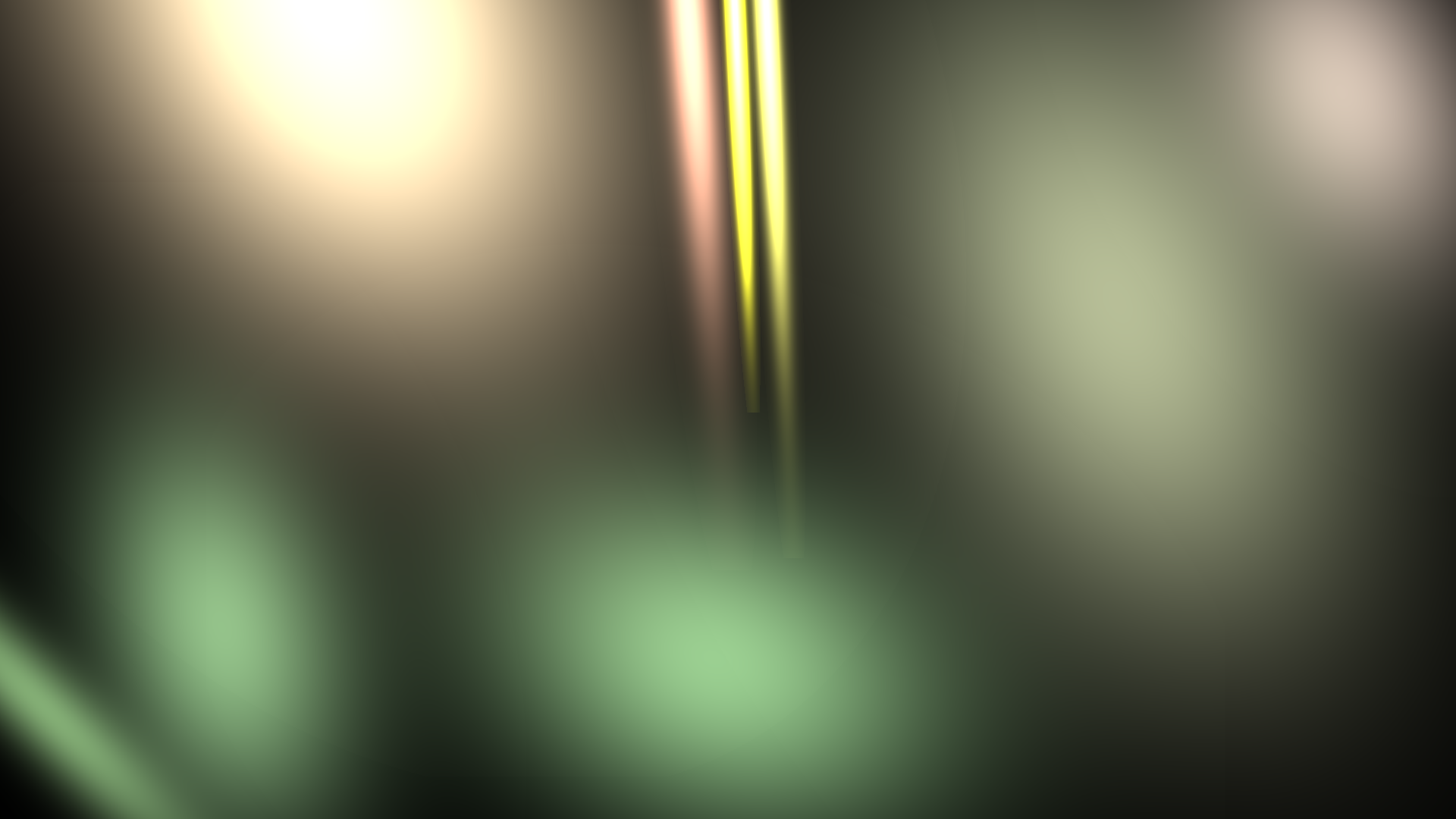}}\hspace{1px}
    \subfloat[$N$ = 100 ($\mathrm{PSNR} = 18.9\,\mathrm{dB}$)]{\includegraphics[width=0.24\textwidth]{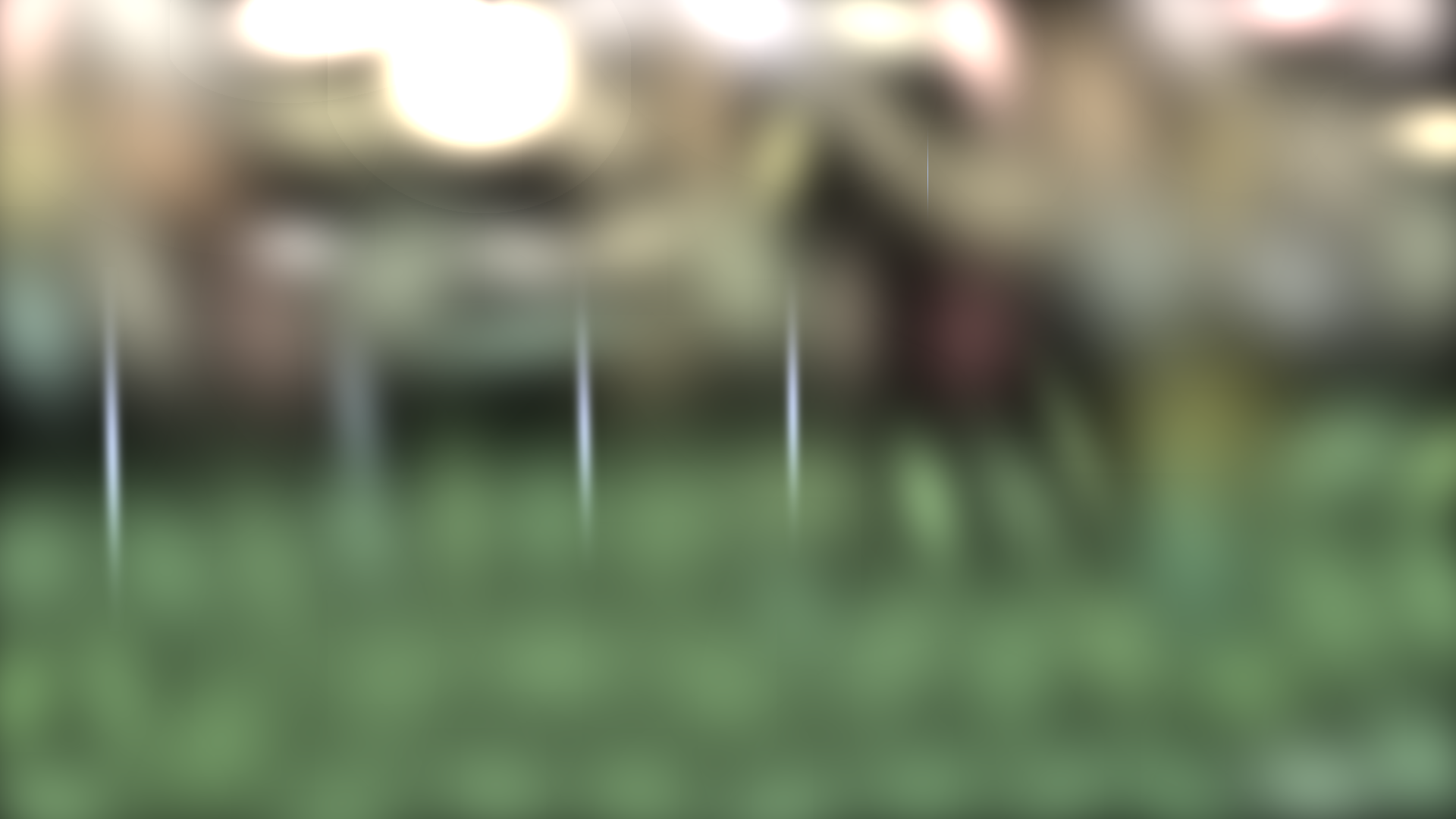}}\hspace{1px}
    \subfloat[$N$ = 1,000 ($\mathrm{PSNR} = 26.5\,\mathrm{dB}$)]{\includegraphics[width=0.24\textwidth]{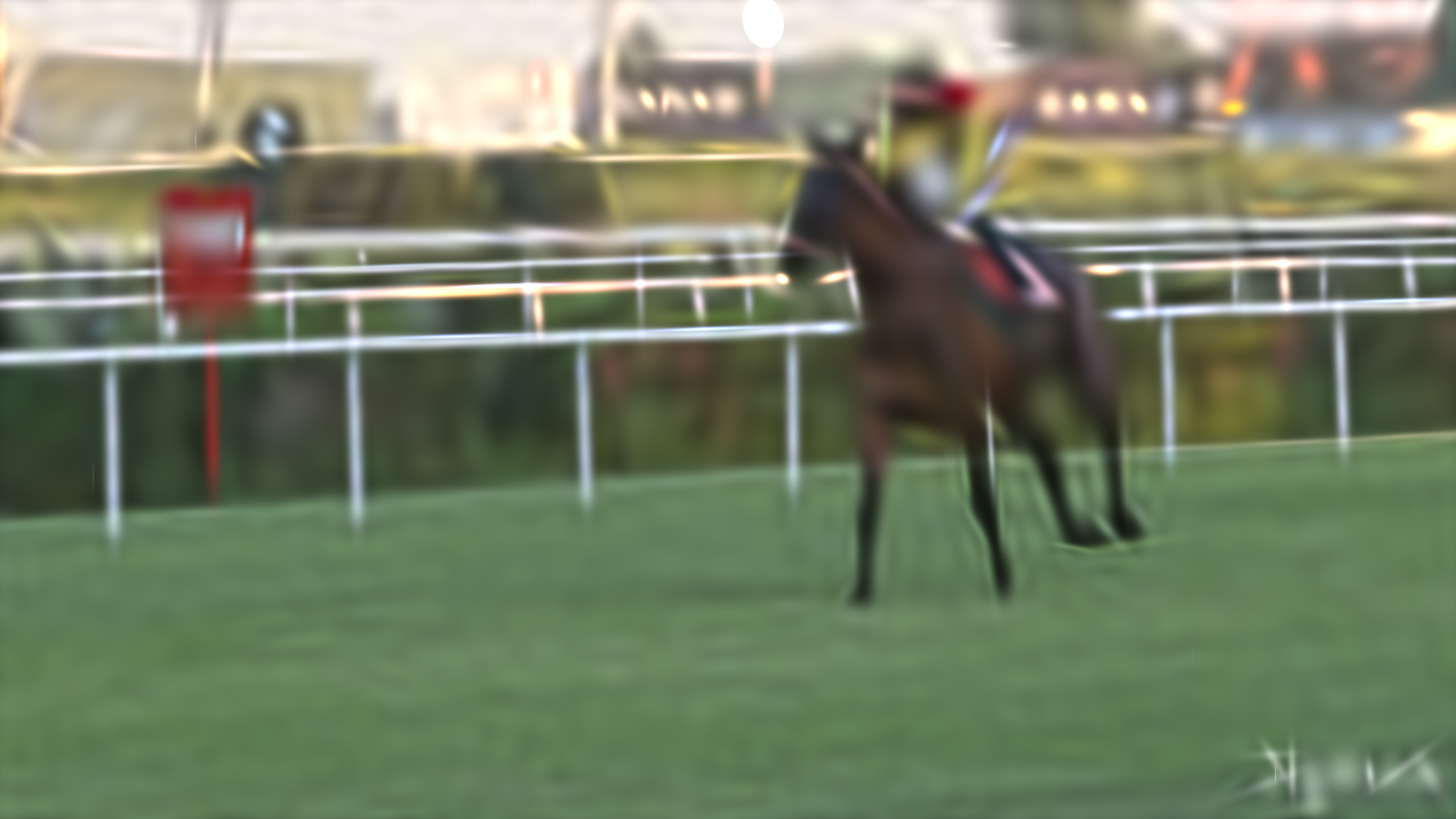}}\hspace{1px}
    \subfloat[$N$ = 10,000 ($\mathrm{PSNR} = 35.2\,\mathrm{dB}$)]{\includegraphics[width=0.24\textwidth]{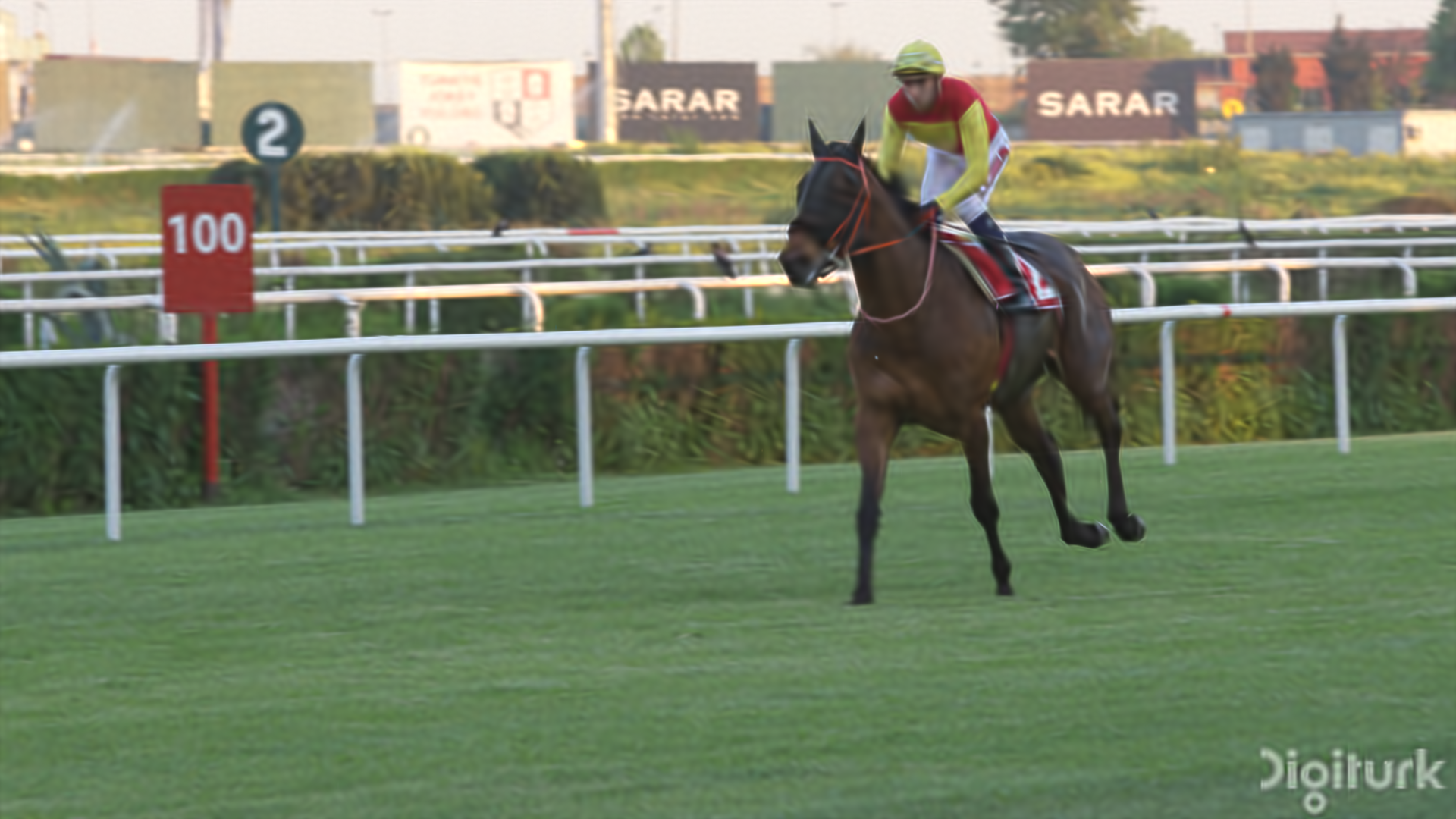}}
    \caption{Using Gaussian splats to model an image.  \rev{The image is taken from the UVG \textit{Jockey} video and has the resolution of $1920\times 1080$.}  The sub-figures show that as the number of Gaussian splats $N$ increases, the learned splats increasingly approximate the image's content.}
    \label{fig:num_gaussians}
\end{figure}

The research community has explored alternative approaches to video compression.
Neural-based approaches (\eg,~\cite{DBLP:conf/cvpr/LiLL23, DBLP:conf/cvpr/LiLL24,shi2024volumetric,DBLP:conf/nips/KwanGZGB23,DBLP:conf/nips/ChenHWRLS21}), driven by recent advances in deep learning, provide a promising alternative by learning a data-driven \textit{implicit} representation that can adapt to diverse video characteristics. 
It remains a challenge, however, for neural-based video encoders to support various ``knobs'' that control trade-offs between encoding quality, compression rate, and encoding time, and to support diverse inputs (\eg, resolutions and the amount of motion) and application contexts (\eg, variable bitrate, low-latency encoding, scalable encoding, and random access). 
As neural-based video encoders adopt an implicit representation, compressed-domain processing is not possible.

Recently, 3D Gaussian Splatting (3DGS)~\cite{DBLP:journals/tog/KerblKLD23} has revolutionized visual data representation for 3D scenes.
Unlike neural-based implicit representations, such as NeRF~\cite{DBLP:journals/cacm/MildenhallSTBRN22}, 3DGS learns explicit Gaussian splats as primitives to represent a 3D scene in high quality.
Using a differentiable, tile-based rasterization framework, 3DGS allows the captured scene to be rendered quickly \rev{on a GPU}.
 
Given the emergence of Gaussian splats, our work in this paper seeks to answer a question: ``\textit{Can Gaussian splats be used as an effective alternative explicit representation for digital video compression?}'' 

Two recent works, VGR~\cite{DBLP:conf/nips/Sun0MLC024} and VeGaS~\cite{DBLP:journals/corr/abs-2411-11024} have applied 3DGS for representing 2D videos.  They are not designed with compression as the main goal, however.  
\rev{On the other hand, 2D Gaussian splatting (2DGS) ~\cite{DBLP:conf/eccv/ZhangGXHWQLGZ24, DBLP:journals/corr/abs-2407-01866} has been proposed as a primitive for image representation.   
Representing visual data using 2D Gaussian splats requires fewer attributes and parameters and thus is more storage efficient than 3DGS.
Rendering is also simpler, as 2DGS does not require depth information. }

\begin{figure}[t]
    \centering
    \subfloat[$t$ = 200 ($\mathrm{PSNR} = 6.1\,\mathrm{dB}$)]{\includegraphics[width=0.24\textwidth]{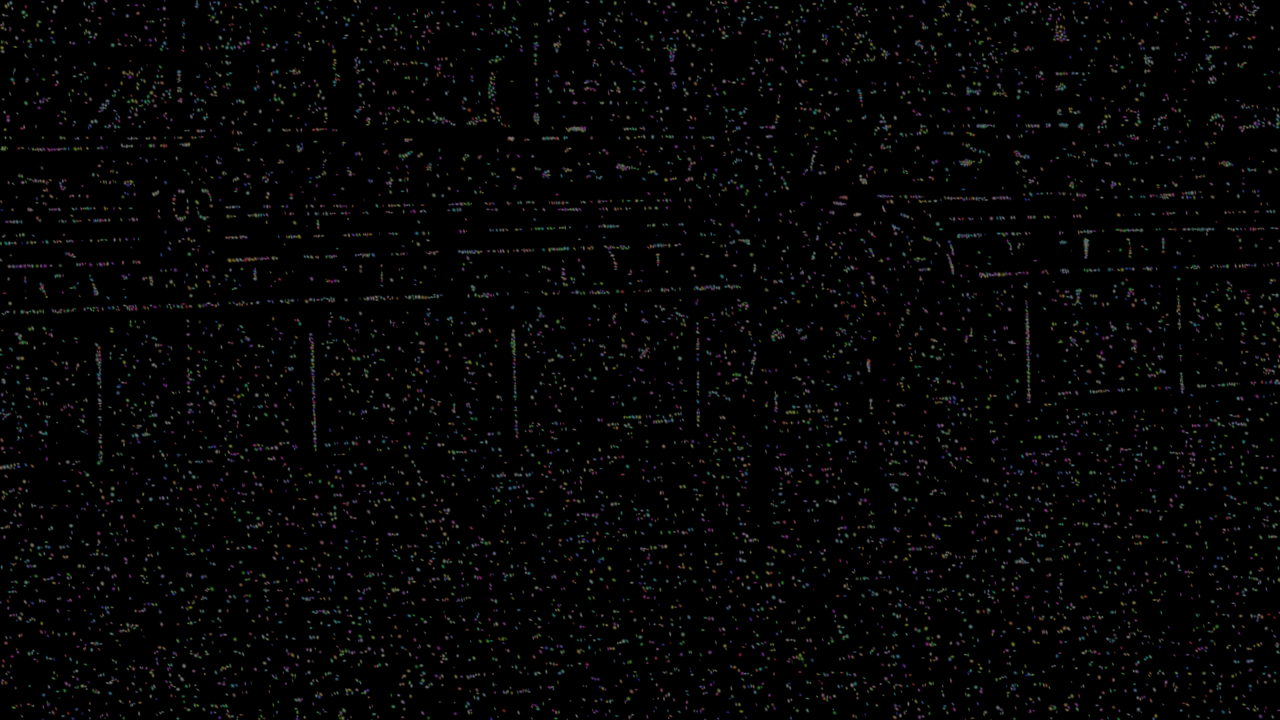}}\hspace{1px}
    \subfloat[$t$ = 800 ($\mathrm{PSNR} = 6.9\,\mathrm{dB}$)]{\includegraphics[width=0.24\textwidth]{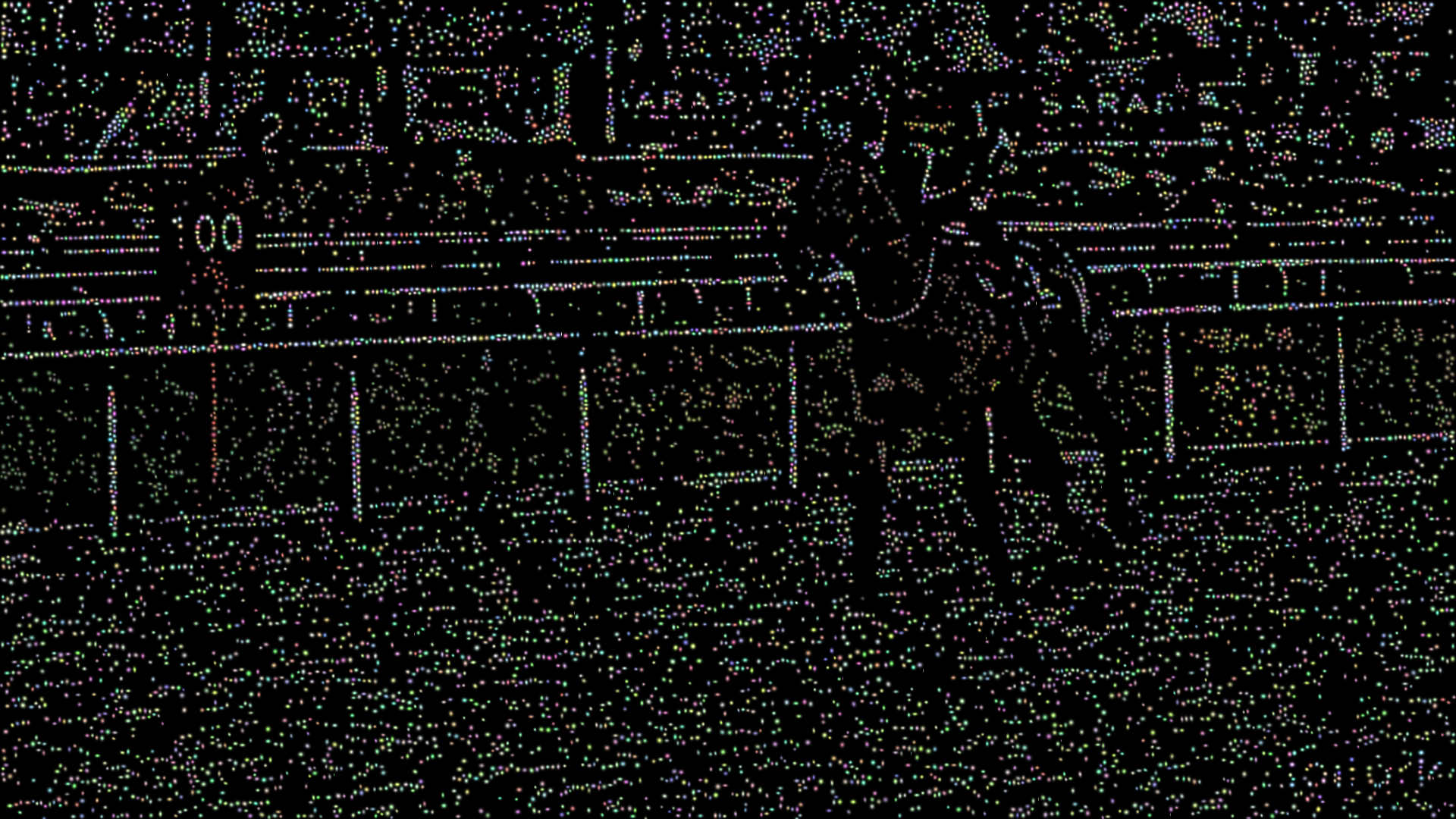}}\hspace{1px}
    \subfloat[$t$ = 3,200 ($\mathrm{PSNR} = 14.7\,\mathrm{dB}$)]{\includegraphics[width=0.24\textwidth]{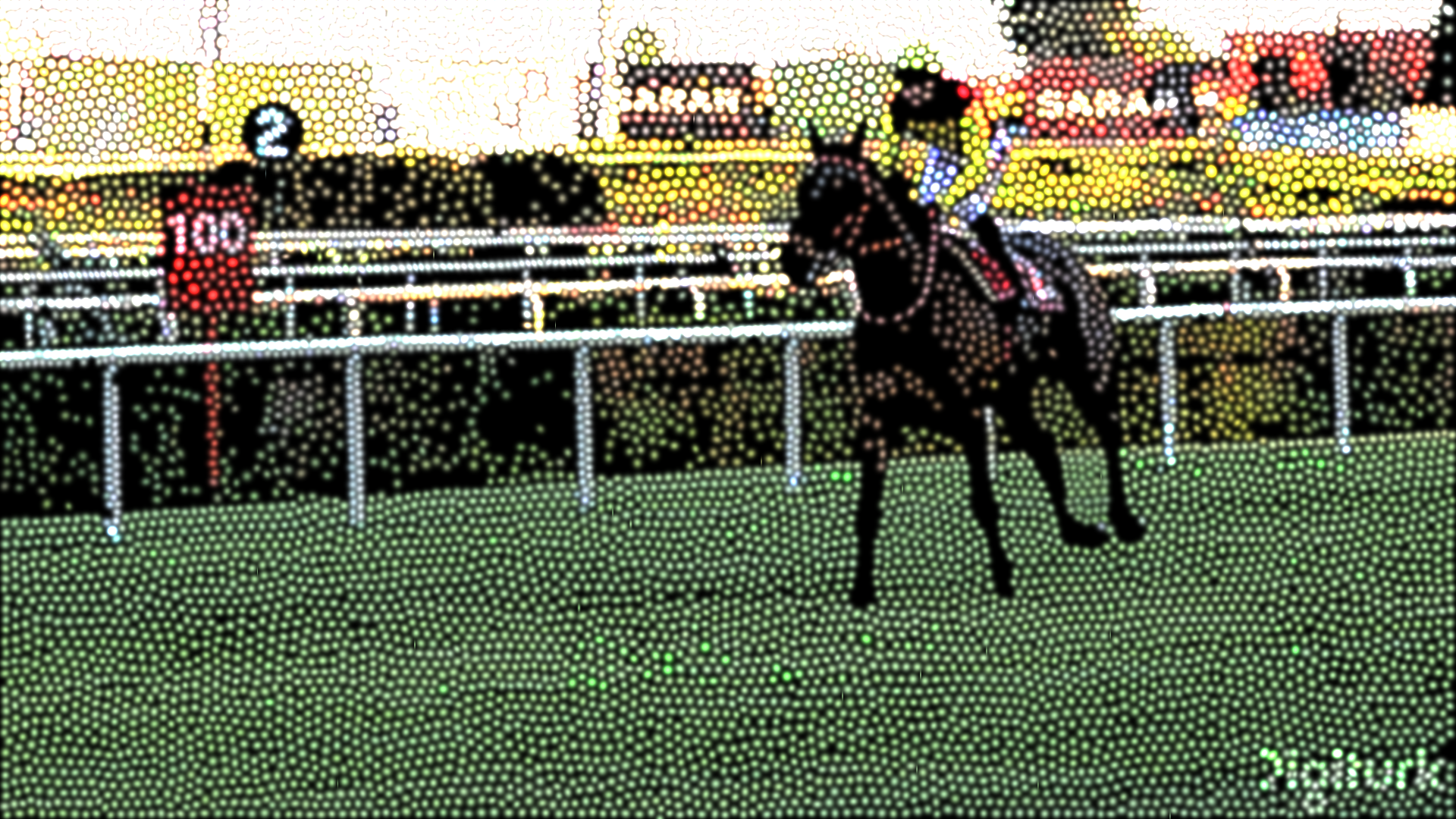}}\hspace{1px}
    \subfloat[$t$ = 12,800 ($\mathrm{PSNR} = 33.7\,\mathrm{dB}$)]{\includegraphics[width=0.24\textwidth]
    {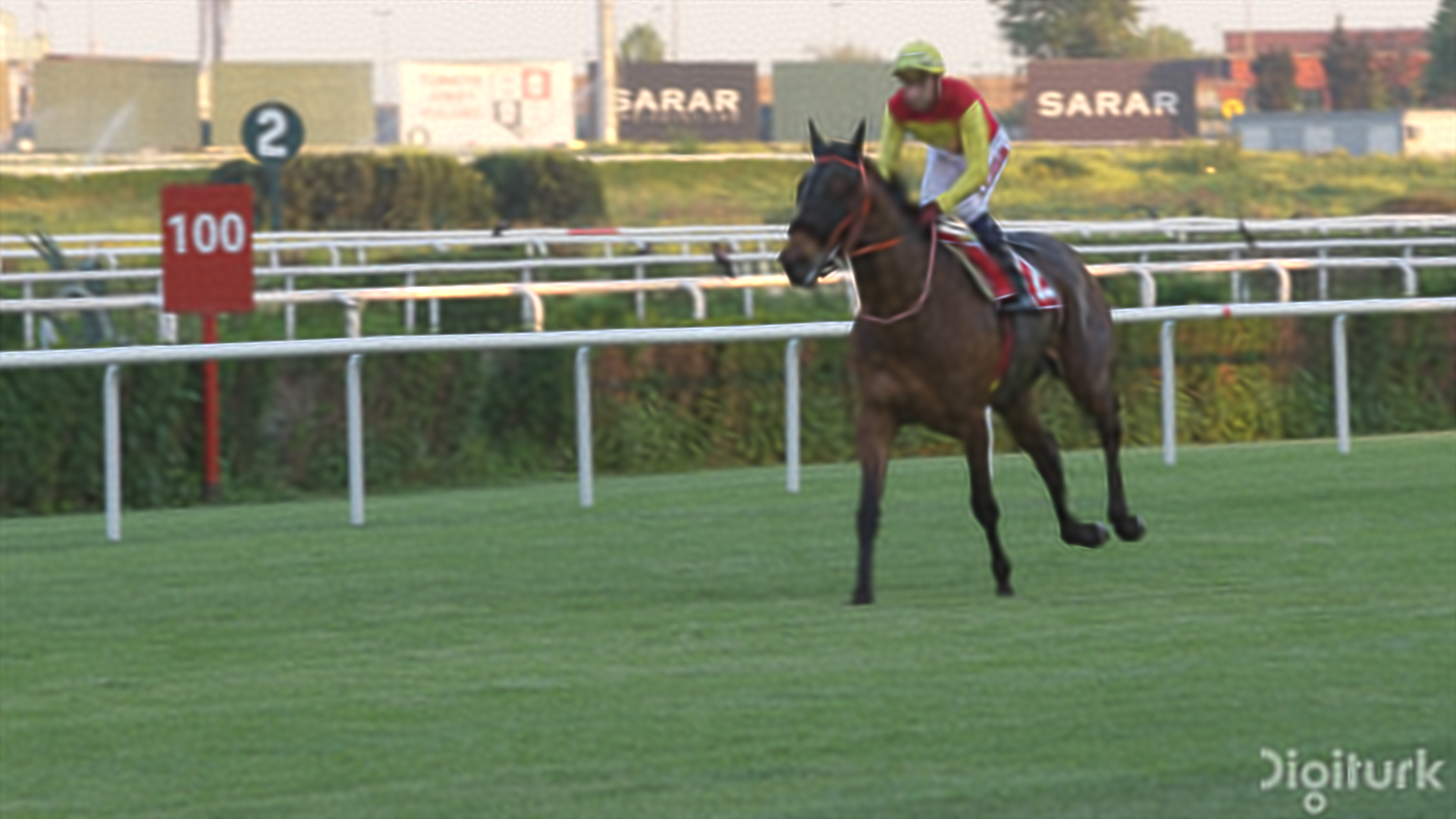}}
    \caption{\rev{Intermediate reconstruction results after $t$ training iterations using 10,000 Gaussian splats on a $1920\times1080$ frame from the UVG \textit{Jockey} video.}}
    \label{fig:num_of_iterations}
\end{figure}

\rev{\Cref{fig:num_gaussians} illustrates how Gaussian splats approximate an image.}
With 10,000 Gaussian splats, we can already represent a perceptually reasonable image. 
\Cref{fig:num_of_iterations} illustrates the training process to optimize the parameters of the Gaussians.
The training process starts with randomly placed Gaussian splats, and iteratively moves, scales, and rotates each Gaussian splat to eventually approximate the given image.

\rev{While 2DGS has been shown to be an effective representation of images, extending it to video representation poses new challenges.}
Particularly, to effectively represent video frames for compression, temporal redundancy and dynamics in the video should be considered.
\rev{Existing work on image representation does not consider the temporal dimension.}

To tackle this challenge, we extend an existing image representation and compression technique, GaussianImage~\cite{DBLP:conf/eccv/ZhangGXHWQLGZ24}, to video, and introduce {\name} (Gaussian Splat-based Video Codec).
Similar to modern video codecs, {\name} categorizes video frames into independent frames and inter-dependent frames.
Unlike existing codecs, however, {\name} treats the concepts of frame dependencies separately during the training and compression.
We briefly introduce the ideas underlying {\name} below.

\rev{{\name} operates in two phases: representation and compression. 
The goal of the representation phase is to find a set of Gaussian splats that best approximate a frame.
The Gaussian splat representation for the first frame in every scene in the video is learned from scratch, initialized with randomly placed Gaussian splats.
We call these frames \textit{anchor frames}, and the Gaussian splats from the anchor frames as \textbf{anchor splats}.}
The anchor frames serve as the priors for the remaining frames.
\rev{For each non-anchor frame, {\name} fine-tunes the Gaussian splats from the previous frame to fit the current frame.
We call the non-anchor frames \textit{propagation frames}, as the Gaussian splats from the anchor frames propagate down to represent these frames.
By reusing Gaussian splats from the previous frame, propagation frames exploit temporal redundancy to accelerate convergence during training.
We refer to splats initialized from the previous frame as \textbf{propagation splats}.}
To cater to video dynamics in propagation frames, {\name} augments each propagation frame with random new anchor splats before fine-tuning.
These augmented splats capture objects with fast motion, or objects that do not appear in the previous frame. 
\rev{{\name} allows rate control to trade off between compression rate and quality through controlling the number of Gaussian splats and the stopping criterion during training: during training, Gaussian splats with low contributions to the frame quality can be pruned to achieve this trade-off. }

When there is a scene change, propagation frames can no longer effectively represent the frames.  To handle this scenario, {\name} monitors the loss values when learning the splats of each frame to determine if a frame should be an anchor frame or a propagation frame, allowing insertion of a new anchor frame when a frame significantly differs from its previous frame.

As anchor frames are initialized from random Gaussian splats, the learned Gaussian splats may not effectively capture the finer details of a scene.  \rev{{\name} supports an optional second pass, where the Gaussian splats from a future propagation frame in the first pass are propagated backward to the preceding anchor frame, to further optimize the splats in the anchor frame (and the following propagation frame).}

The processes above describe how to \textit{learn to represent} video frames with 2D Gaussian splats.
With the representations, {\name} then proceeds to the second phase, the compression phase.  {\name} learns the scaling factor, the offset factor, and the color codebook to encode the Gaussian splats and adjusts the Gaussian splats according to the quantized parameters.
This process is similar to that from GaussianImage~\cite{DBLP:conf/eccv/ZhangGXHWQLGZ24} for anchor frames.
\rev{For propagation frames, {\name} compresses only the changes in Gaussian parameters relative to the previous frame, reducing storage requirements.}

In the context of encoding, we adopt the term intra-coded frame (I-frame) and predictive-coded frame (P-frame) from standard coding.
\rev{Anchor frames are coded as I-frames using absolute Gaussian parameters, while propagation frames are typically coded as P-frames using parameter residuals.
For better error robustness and to prevent quantization errors from propagating, periodically, a propagation frame is intra-coded as an I-frame.
The compressed version requires only 56 bits for each intra-coded splat and 45 bits for each predictive-coded splat.
An additional 216 bytes and 408 bytes are needed to encode color codebooks and other parameters for each I-frame and for each P-frame, respectively.}

This paper extends an earlier version of our work~\cite{DBLP:conf/nossdav/WangSO25} in \rev{three aspects}:
\begin{itemize}
    \item \rev{We introduce a two-pass encoding strategy that leverages future frames to enhance anchor frames, achieving over 0.5 dB PSNR gain on the most challenging UVG sequences.}
    \item \rev{We introduce residual-based quantization with separate quantization of Gaussian parameter updates, reducing compressed size by over 34\% while maintaining comparable or better quality than our earlier version.}
    \item \rev{We now decouple the notion of independent and inter-dependent frames for the representation phase and compression phase, allowing inter-dependent frames to exploit priors and find better Gaussian splat representations during training, while avoiding error propagation due to quantization errors during compression.}
\end{itemize}
Furthermore, we extend our evaluation to the full set of 1080p test videos in the UVG dataset, and conduct additional ablation experiments, with more detailed analyses, to further validate the overall effectiveness of GSVC and the benefits of these extensions.

The main contributions of {\name} are summarized as follows:
\begin{itemize}
\item \rev{We demonstrate the effectiveness of 2D Gaussian splats as a new primitive for video representation and compression.}
\item  \rev{We address the challenges of learning Gaussian splats for video representation and compression using several techniques: incremental learning, two-pass encoding, Gaussian splat pruning, Gaussian splat augmentation, and key-frame identification.  With these techniques, our approach enables a compact representation while effectively capturing newly appearing objects, large deformations, and scene transitions.}
\item We improve the 2D Gaussian video compression pipeline by explicitly tracking Gaussians’ deltas across frames and quantizing different temporal changes (parameter updates and Gaussian insertions/removals) separately, leading to a more compact video bitstream. 
\item \rev{We demonstrate that {\name} achieves rate-distortion performance comparable to open-source implementations of state-of-the-art video coding standards on the UVG and HEVC Class B datasets.}
\end{itemize}

\section{Related Work}
\subsection{Video Compression}

Signal processing techniques, particularly frequency-domain representations, form the foundation of modern video compression standards such as AVC~\cite{DBLP:journals/tcsv/WiegandSBL03}, HEVC~\cite{DBLP:journals/tcsv/SullivanOHW12}, and VVC~\cite{DBLP:conf/vcip/Sullivan20}.
These standards rely on hand-crafted techniques such as block-based motion compensation and transform coding to achieve efficient compression~\cite{DBLP:conf/cvpr/LuO0ZCG19}.
Despite decades of advances in standardized video coding, researchers continue to explore alternative approaches to video compression.

Recent neural-based approaches provide a promising alternative to traditional codecs~\cite{DBLP:conf/mm/Li0022,DBLP:conf/cvpr/LiLL23,shi2024volumetric,DBLP:conf/nips/KwanGZGB23,DBLP:conf/nips/ChenHWRLS21,DBLP:journals/corr/abs-2501-01681}.
\rev{These methods leverage deep learning to optimize compression pipelines and learn data-driven representations that adapt to different video characteristics~\cite{DBLP:conf/cvpr/LuO0ZCG19}.
However, implicit neural representations have several limitations.
First, decoding requires neural network inference~\cite{DBLP:journals/corr/abs-2501-01681}, which can be computationally expensive.
Second, storing compressed content in implicit latent features or network parameters can complicate spatial random access, such as decoding an arbitrary image region.
Third, implicit representations limit direct editability, as operations such as spatial scaling, appearance modification, geometry adjustment, or local editing may require additional processing or re-encoding~\cite{DBLP:journals/tog/KastenOWD21,DBLP:journals/corr/abs-2307-07663,DBLP:conf/mm/Lu0GPXMXW24}.
Another direction integrates neural networks into conventional video coding frameworks.
For example, MFRNet~\cite{ma2021mfrnet} employs CNN-based post-processing and in-loop filtering for HEVC and VVC, while Man et al.~\cite{man2025content} introduce a content-aware neural in-loop filter into VVC with adjustable computational complexity.
In contrast to implicit representations, {\name} represents videos using explicit 2D Gaussian splats, where each primitive directly stores its position, shape, and color parameters.}

\rev{Scene-change detection is another important component for managing temporal dependencies in video coding.
Conventional approaches detect scene transitions from image-domain differences, using features such as HSV color histograms~\cite{DBLP:conf/iotdcc/SunZ17} or local descriptors such as ORB~\cite{DBLP:conf/iccv/RubleeRKB11}, while learning-based methods such as TransNet V2~\cite{soucek2024transnetv2} directly learn to identify shot transitions from video frames.
Different from these image-domain approaches, {\name} performs scene-change detection in the Gaussian representation domain by evaluating whether the propagated Gaussian representation remains suitable for the current frame.}

\subsection{Gaussian Splatting for Video}
\rev{3D Gaussian Splatting (3DGS)~\cite{DBLP:journals/tog/KerblKLD23} was initially proposed for static 3D scene reconstruction, where a set of 3D Gaussian splats with learnable positions, covariances, and colors provides an explicit, differentiable representation of the underlying radiance field.
Recent studies have extended 3DGS to video representation.
For instance, VGR~\cite{DBLP:conf/nips/Sun0MLC024} represents each video frame using 3D Gaussian splats, treating it as a static 3D scene, while VeGaS~\cite{DBLP:journals/corr/abs-2411-11024} treats video sequences as 3D data with time as the third dimension and demonstrates various video processing operations, such as interpolation, spatial scaling, and temporal manipulation, within this Gaussian domain.
However, they are not specifically designed for compression.
Meanwhile, they inherit the high parameter count and computational overhead of 3DGS, which requires depth sorting and alpha blending during rendering.
Modeling complex temporal dynamics is also challenging, as large motions and newly appearing objects require substantial changes to the Gaussian representation across frames~\cite{DBLP:conf/nips/Sun0MLC024}.}

More recently, 2D Gaussian splats have been introduced for image representation and compression.
GaussianImage~\cite{DBLP:conf/eccv/ZhangGXHWQLGZ24} shows that a single image can be compactly reconstructed by a set of 2D Gaussians, and LIG~\cite{DBLP:conf/aaai/ZhuLCZJWY25} further adopts a progressive two-level Gaussian structure to improve reconstruction quality, especially for large-resolution images.
\rev{Compared with 3DGS, 2DGS uses far fewer parameters per splat (e.g., 8 instead of 59 in 3DGS) and enables more efficient rendering without requiring depth estimation or sorting.}
Extending 2D Gaussian splatting from static images to videos, however, is challenging: unlike single images, videos exhibit substantial temporal dynamics, including motion, lighting changes, and occlusions, and a static 2DGS representation would ignore temporal redundancy and lead to inefficient compression.
To address these challenges, 
our earlier work GSVC~\cite{DBLP:conf/nossdav/WangSO25} first introduced 2D Gaussian splats as a new primitive for video representation and compression and achieved competitive rate–distortion performance compared to modern codecs such as AV1 and VVC.
In this journal extension, we further enhance GSVC in terms of both representation and compression.

\section{Video Representation through 2D Gaussian Splats}
We begin by formulating the problem in \Cref{sec:formulation}, followed by an in-depth discussion on the structure and rendering of 2D Gaussian splats in \Cref{sec:2dgaussian}.
We then present the mechanism for Gaussian splat propagation in~\Cref{sec:prediction} and two-pass training in~\Cref{sec:enhancement}.
Then, in \Cref{sec:pruning}, we describe how {\name} removes Gaussian splats to reduce the size with minimal impact on the quality.
\rev{In \Cref{sec:dynamic}, we introduce the Gaussian splat augmentation process, designed to add new Gaussian splats, accommodating large deformations and abrupt objects.
\Cref{sec:keyframe} details our approach to dynamically select anchor frames to prevent unrelated scene priors from affecting model performance.
Finally, we integrate all components to illustrate the complete end-to-end representation training pipeline in \Cref{sec:together}.}

\subsection{Problem Formulation \label{sec:formulation}}
\rev{We first introduce the notation for our video encoding pipeline.
Let $\boldsymbol{\mathcal{F}} = \{f_1, f_2, \dots, f_T\}$ denote a video with $T$ frames.}
Our objective is to encode it as $\widehat{\boldsymbol{\mathcal{F}}} = \{\widehat{f}_1, \widehat{f}_2, \dots, \widehat{f}_T\}$.
In $\widehat{\boldsymbol{\mathcal{F}}}$, each frame $\widehat{f}_t$ is represented by a set of 2D Gaussian splats $\boldsymbol{\mathcal{G}}_t = \{\boldsymbol{G}_{t,1}, \boldsymbol{G}_{t,2}, \dots, \boldsymbol{G}_{t,N_t}\}$, where $N_t$ is the number of Gaussian splats in $f_t$.
\rev{The number of Gaussian splats $N_t$ serves as a rate-control parameter for trading off quality and size.
For ease of exposition, we set $N_t=N$ for all frames in both our description and implementation.}
One could easily adjust the $N_t$ according to the complexity of the scene on a per-frame basis.
The entire video can be expressed as:
$\boldsymbol{\mathcal{G}} = \{\boldsymbol{\mathcal{G}}_{1}, \boldsymbol{\mathcal{G}}_{2}, \dots, \boldsymbol{\mathcal{G}}_{T}\}$.

\subsection{2D Gaussian Image Representation \label{sec:2dgaussian}}
To achieve an efficient Gaussian representation, it is essential to maintain a compact and flexible Gaussian splat structure.
\rev{Following prior 2DGS image representations~\cite{DBLP:conf/eccv/ZhangGXHWQLGZ24, DBLP:journals/corr/abs-2407-01866}, we adopt a similar parameterization and rendering approach.}
Each Gaussian splat is characterized by three attributes, comprising a total of eight parameters: position $\boldsymbol{\mu} \in \mathbb{R}^2$, weighted color $\boldsymbol{c}' \in \mathbb{R}^3$, and Cholesky vector $\boldsymbol{\ell} \in \mathbb{R}^3$.
Specifically, the weighted color $\boldsymbol{c}'$ is formed by integrating the opacity $o \in \mathbb{R}$ into the color coefficients $\boldsymbol{c} \in \mathbb{R}^3$.
\rev{The Cholesky vector $\boldsymbol{\ell} = \{\ell_{0}, \ell_{1}, \ell_{2}\}$ represents the lower triangular elements of the matrix $\boldsymbol{L}$}, which describes the covariance matrix $\boldsymbol{\Sigma} \in \mathbb{R}^{2 \times 2}$ using the Cholesky decomposition~\cite{higham2009cholesky}:
$\boldsymbol{\Sigma} = \boldsymbol{L}\boldsymbol{L}^T$.
This decomposition ensures the covariance matrix remains positive semi-definite during optimization, preventing invalid parameter updates.

At a high level, the training process works as follows (the details and complete pipeline will be given in~\Cref{sec:together}).  Given an input RGB image $F_t$, we repeatedly optimize the number of Gaussians and their parameters so that the set of Gaussian splats, $G_t$, when rendered, approximates $F_t$ as much as possible.

For rendering, the color $\boldsymbol{C}_{i}$ at the $i$-th pixel is computed using an accumulated summation mechanism based on $\boldsymbol{c}'$: 
\begin{equation} 
\label{rendering}
    \begin{aligned}
        \boldsymbol{C}_{i} & = \sum_{n \in \mathcal{N}} \boldsymbol{c}_n' \cdot \exp(-\boldsymbol{\sigma_n}), 
        \\
        \boldsymbol{\sigma_n} & = \frac{1}{2} \boldsymbol{d}_n^T \boldsymbol{\Sigma}^{-1} \boldsymbol{d}_n,
    \end{aligned}
\end{equation}
\rev{where $\mathcal{N}$ denotes the set of Gaussian splats covering the $i$-th pixel, and  $\boldsymbol{d_n}$ denotes the distance between the pixel center and the projected Gaussian center.}

\subsection{Gaussian Splat Propagation\label{sec:prediction}}
\rev{A naive approach to representing a video with Gaussian splats is to represent each frame independently.}
Doing so is analogous to encoding a video as a Motion JPEG or an I-frame-only MPEG video.  However, we can exploit temporal redundancy in a video to improve encoding (\ie, training) speed and encoding efficiency.  

{\name} trains the first frame in a video sequence in a way similar to that of GaussianImage, independently of any previous frames.  The set of Gaussian splats is initialized from random parameters, and then trained to approximate the image.   We call such a frame an \textit{anchor frame}, and the splats from these frames as \textit{anchor splats}.

For non-anchor frames, {\name} uses the Gaussian splats from the previous frame as priors. It first propagates the trained Gaussian splats from the previous frame as the initial set of Gaussian splats. 
The parameters of these Gaussian splats are thus fine-tuned to fit the current frame.  Only the changes between the parameters of each Gaussian splat are stored.  The changes in the position of Gaussian splats are analogous to motion vectors; changes to the other parameters are analogous to the residuals. \rev{We refer to these frames as \textit{propagation frames} and their splats as \textit{propagation splats}.
We extend our notation with a superscript to denote a frame as either an anchor frame ($\boldsymbol{\mathcal{G}}_{t}^{a}$) or propagation frame ($\boldsymbol{\mathcal{G}}_{t}^{p}$).}

\rev{Note here that, while the notions of anchor frames and propagation frames are similar to those of intra-coded frames (I-frames) and predicted frames (P-frames), we avoid using the same term.  {\name} decouples the notion of interdependent frames during training and during compression.  During training, inter-dependent frames (\ie, propagation frames) are used to improve the quality of the representation; during compression, these frames may still be independently coded with intra-coding to prevent error propagation.}

\subsection{\rev{Two-pass Encoding for Anchor Frames}\label{sec:enhancement}}
\begin{figure}[t]
    \centering
    \subfloat[Gaussian Splat Propagation]{\parbox{.49\textwidth}{%
        \includegraphics[width=0.48\textwidth]{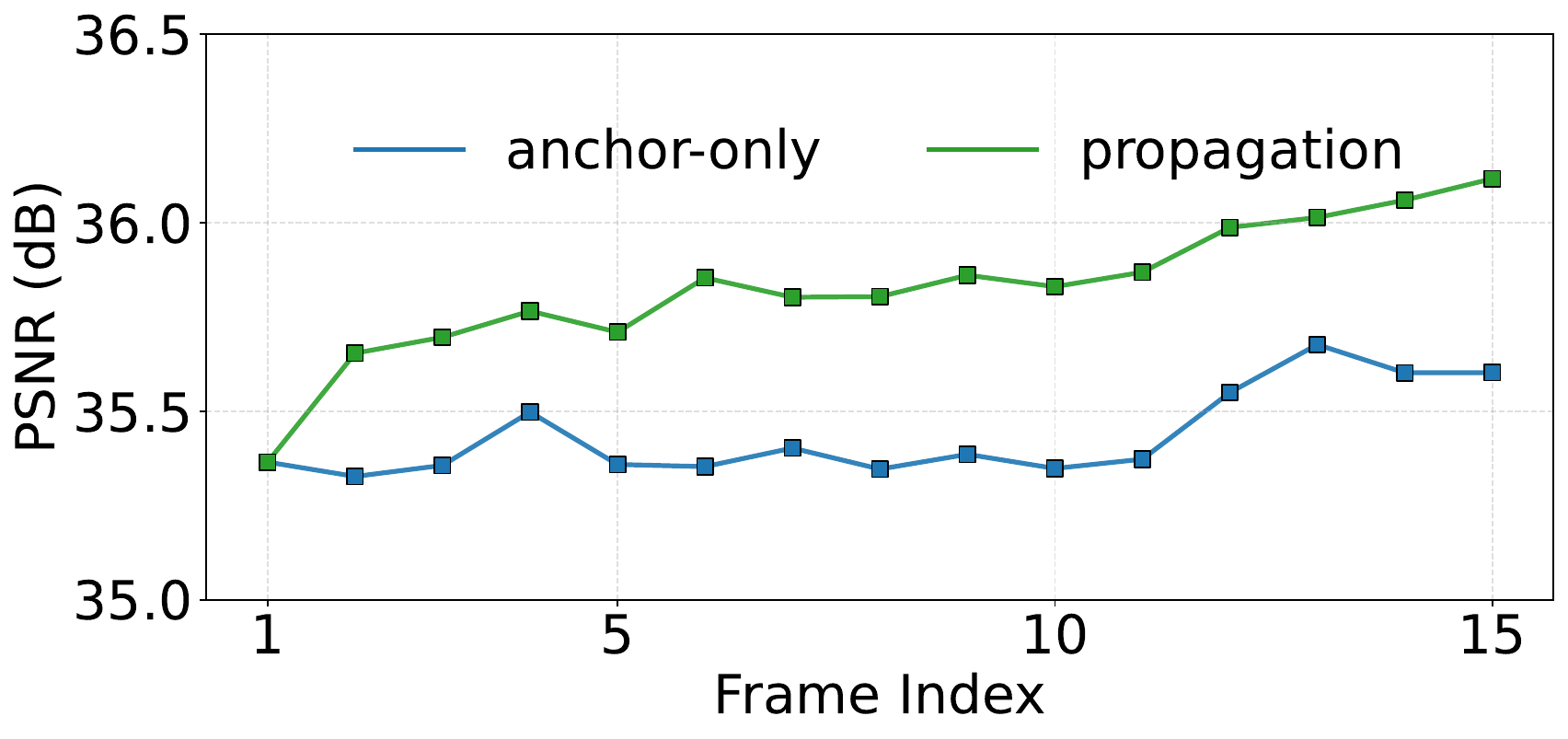}\label{fig:I_vs_IP}
    }}\hspace{1mm}
    \subfloat[Two-pass Encoding]{\parbox{.49\textwidth}{%
        \includegraphics[width=0.48\textwidth]{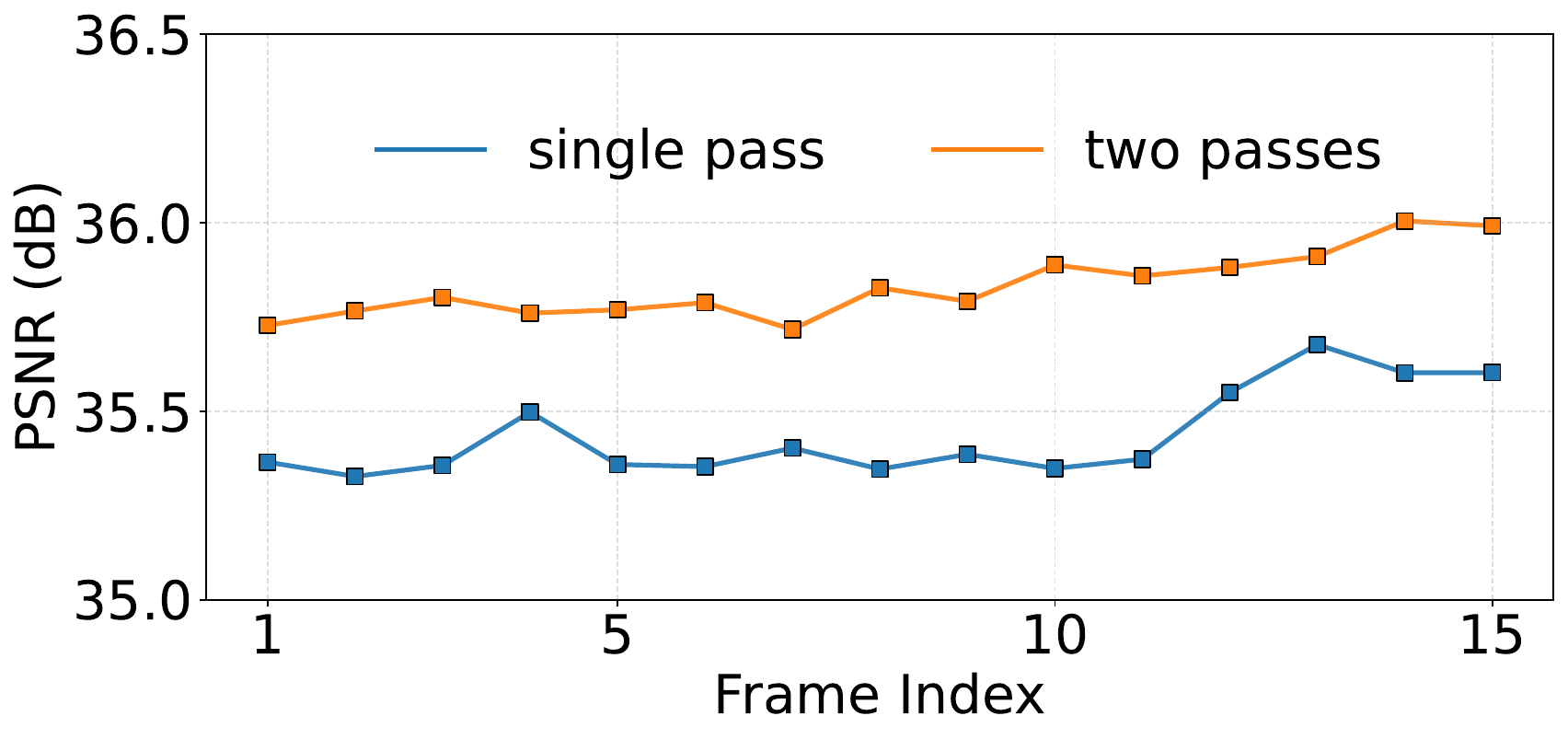}\label{fig:I_vs_I*}
    }}
    \caption{
    PSNR across frames 1–15 on Jockey with 10k Gaussian splats for (a) anchor-only reconstruction versus frame-prediction reconstruction (first frame as anchor frame and remaining frames as propagation frames), and (b) impact of two-pass training, comparing the original anchor frame (single pass) with the refined anchor frame obtained through two-pass training (two passes).  
     }
    \label{fig:I_vs_IP_vs_I*}
\end{figure}

\rev{We observe that propagation frames achieve higher reconstruction quality than anchor frames, with quality generally improving as propagation proceeds.}
\Cref{fig:I_vs_IP} illustrates this phenomenon, where the first frame, an anchor frame, is the lowest quality frame when we encode with both anchor frames and propagation frames.  Using propagation frames incrementally improves the quality.  \rev{With only anchor frames, the frame quality remains mostly constant.}

\rev{We first emphasize that this progressive improvement is \emph{not} an artifact of incomplete optimization but results from more effective initialization in propagation frames.}
For each frame, we train the corresponding Gaussian splats until convergence using the same convergence threshold across the entire sequence, ensuring that future frames do not simply receive more training than previous frames or anchor frames.
Given this per-frame convergence setting, the remaining performance gap can be naturally attributed to the initialization: each frame is initialized from previously trained frames, whose Gaussian parameters have already been refined into better and more temporally consistent representations of the underlying scene.
\rev{As training progresses, these inherited Gaussian splats provide increasingly accurate priors, enabling subsequent frames to converge to higher-quality optima with improved reconstruction fidelity.}

\rev{To exploit the priors from propagation frames, {\name} optionally performs a second pass to refine anchor frames using information from future propagation frames.
In the first pass, each anchor frame is randomly initialized and trained from scratch.}
In the second pass, we train the next $k$ frames as propagation frames and train them sequentially through Gaussian propagation, where $k$ denotes the number of future frames used for refinement ($k=1$ in our experiments and in~\Cref{fig:I_vs_I*}).
After that, we re-initialize the anchor frame with the Gaussian splats from the $k$-th frame and train the anchor frame again. 
This two-pass strategy effectively lets the anchor frame inherit the accumulated temporal knowledge from future frames, leading to more stable color statistics, finer geometric consistency, and smoother temporal transitions.
\rev{Note that the Gaussian splats in the anchor frame remain as anchor splats.}

\Cref{fig:I_vs_I*} illustrates the impact of the two-pass training on anchor frames. In this experiment, we use a segment of 16 frames and report the refined reconstruction quality of the first 15 frames after the two-pass procedure, trained as an \textit{anchor-only sequence}.  \rev{As shown in~\Cref{fig:I_vs_I*}, using a two-pass strategy consistently achieves higher PSNR (over 0.5 dB) than optimizing with only a single pass. }

\rev{After two-pass encoding, the re-optimized anchor frame is used as a new source for Gaussian splat propagation. {\name} retrains subsequent propagation frames propagated from this anchor frame, so that their Gaussian splats inherit the improved color statistics and geometry, leading to better frame quality and more consistent temporal transitions.
Notably, two-pass encoding incurs additional encoder-side computation, as the look-ahead frames are optimized before the final forward propagation.
However, GPU memory overhead remains small because these frames are optimized consecutively and temporary buffers can be reused.
Since only the final refined anchor and subsequent propagation frames are encoded, the temporary representations introduce no additional bitstream overhead. We further quantify the encoding-latency and GPU memory overhead under different look-ahead lengths in~\Cref{sec:results_evaluation}.}

\subsection{Gaussian Splat Pruning \label{sec:pruning}}

\begin{figure}[t]
    \centering
    \begin{minipage}[b]{\textwidth}\centering
        \subfloat[Baseline ($f_1$)]{\includegraphics[width=0.48\textwidth]{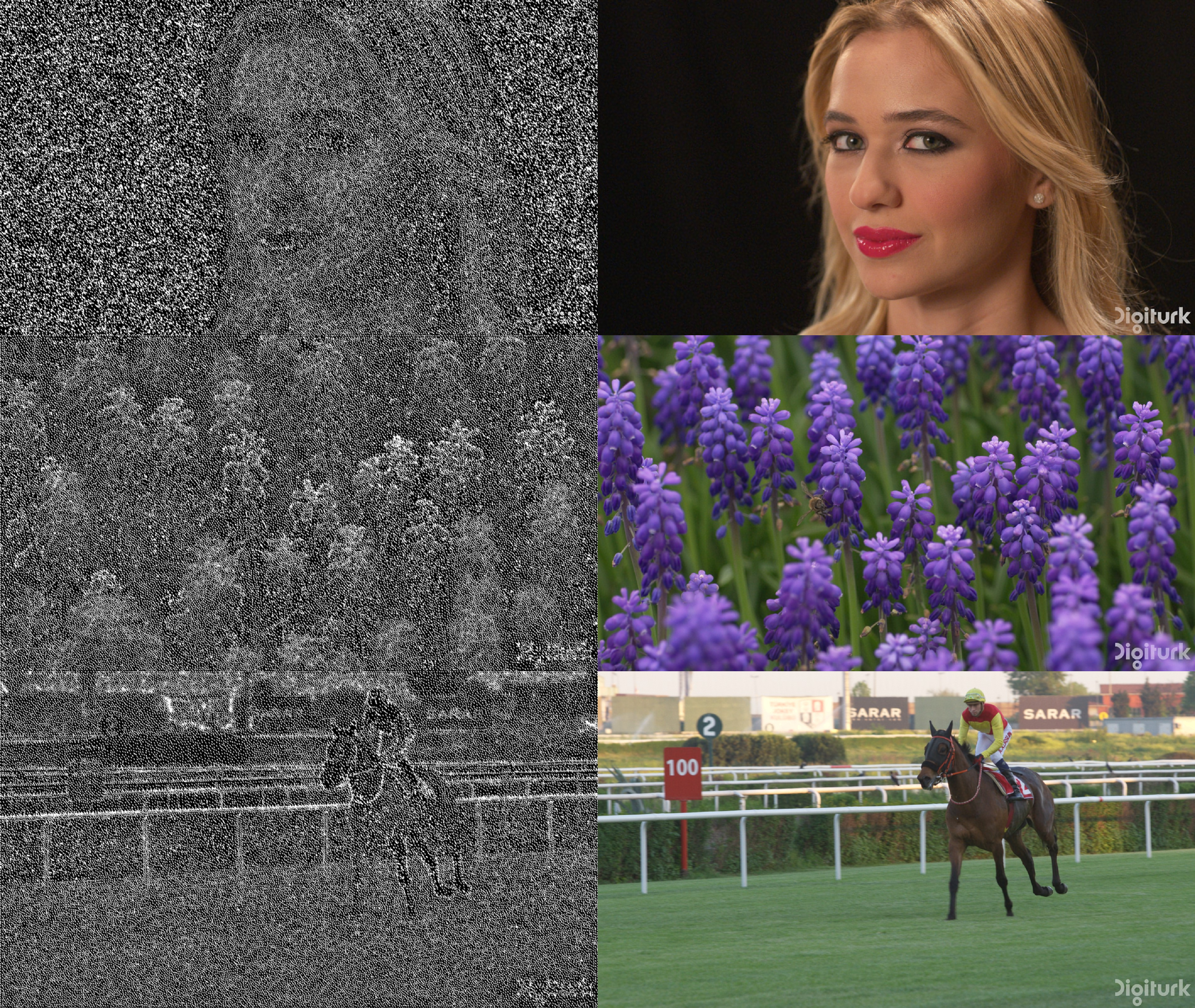}\label{BaselineUVG}}\hspace{1px}
        \subfloat[GSP ($f_1$)]{\includegraphics[width=0.48\textwidth]{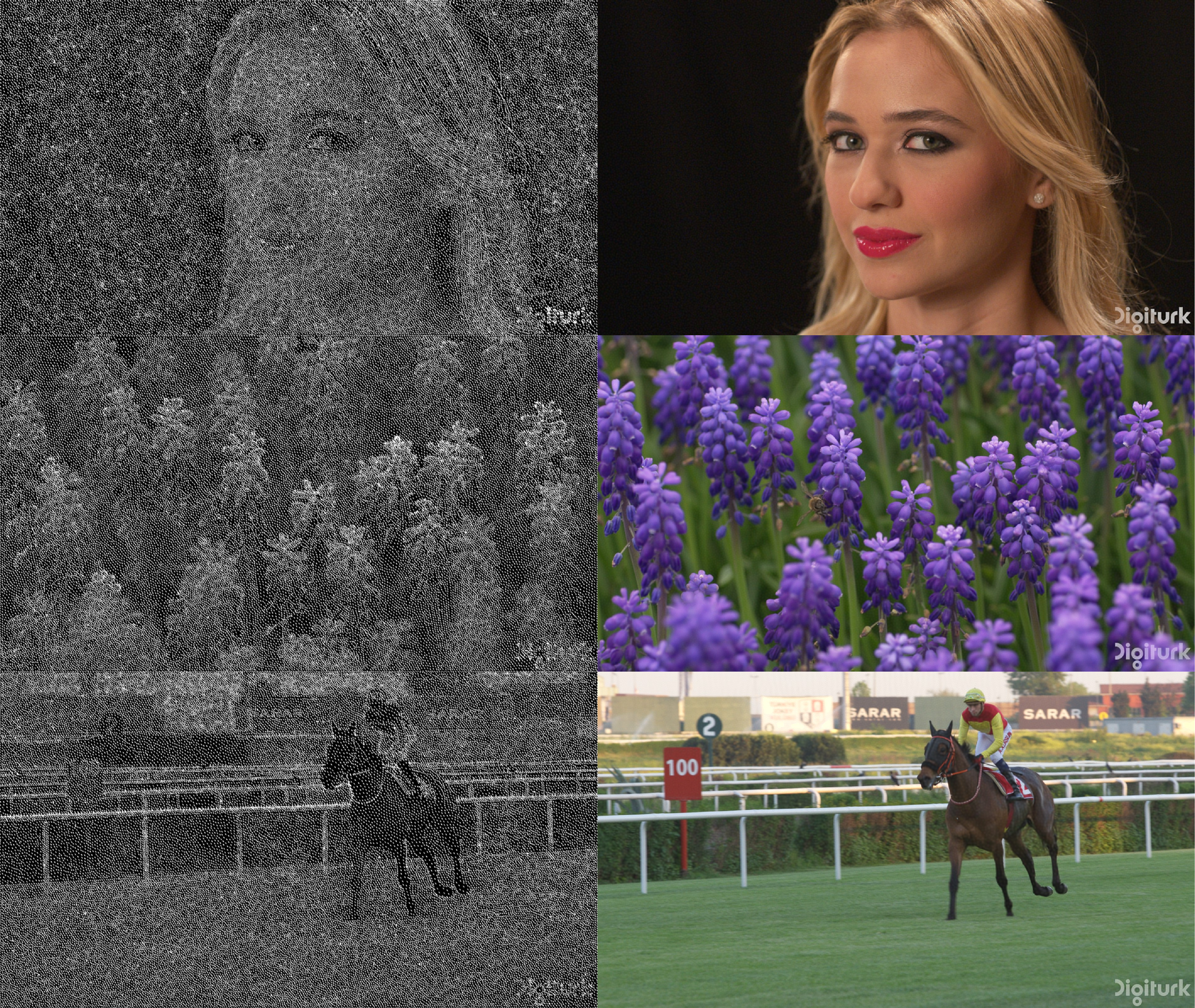}\label{GSPUVG}}
    \end{minipage}
    \caption{\rev{Gaussian Splat Center Distributions ($45k$) and Corresponding Renderings on $f_1$ of \textit{Beauty}, \textit{HoneyBee}, and \textit{Jockey}.}}
    \label{fig:pruning}
\end{figure}

A key step in video encoding is to tune the trade-off between the encoding rate and the quality.  For constant bitrate encoding, the encoder attempts to maintain a constant bitrate throughout the video but allows the quality to vary.  In our context, constant bitrate is achieved by keeping the same number of Gaussian splats $N$ for every frame.  On the other hand, for constant quality encoding, the encoder keeps the frame quality constant but allows the bitrate to vary.  In {\name}, this corresponds to removing Gaussian splats during the training process, while keeping the frame quality above a threshold. 

For either approach above, a critical step is to remove Gaussian splats with minimal impact on the image quality.  
\rev{To this end, we introduce Gaussian Splat Pruning (GSP).
Specifically, a learnable parameter $\boldsymbol{w}$ is introduced to modulate the contribution of each Gaussian splat through its weighted color $\boldsymbol{c}'$.}
Accordingly, the rendering formula in~\Cref{rendering} is modified as
\begin{equation} 
\label{renderingW}
    \boldsymbol{C}_{i} = \sum_{n \in \mathcal{N}} 
    \boldsymbol{w}_n \boldsymbol{c}_n' \cdot \exp(-\boldsymbol{\sigma}_n).
\end{equation}

\rev{To account for both color contribution and spatial coverage, we define the importance score of the $n$-th Gaussian as }
\begin{equation} 
\label{eq:importance}
s_n = \|\boldsymbol{w}_n\|_2\left|\ell_{0,n}\ell_{2,n}\right|,
\end{equation}
\rev{where $\ell_{0,n}$ and $\ell_{2,n}$ are the diagonal elements of its Cholesky factor, and $|\ell_{0,n}\ell_{2,n}|$ is proportional to the Gaussian's spatial footprint.}
Gaussians with lower importance scores are pruned first, reducing the number of splats while minimizing the impact on reconstruction quality.
$N_{\text{prune}}$ provides a ``knob'' to trade off size and quality.

Note that $w$ is not propagated to subsequent propagation frames $\boldsymbol{\mathcal{G}}_{t+1}^{P}$ to avoid increasing the model size. Instead, the product $\boldsymbol{w}\boldsymbol{c}'$ in $\boldsymbol{\mathcal{G}}_{t}^{*}$ is used to initialize the weighted color coefficient $\boldsymbol{c}'$ in $\boldsymbol{\mathcal{G}}_{t+1}^{P}$.

\rev{\Cref{fig:pruning} compares training 45,000 Gaussian splats directly with GSP initialized from 50,000 splats and pruned to 45,000.}
\rev{The results show that the remaining Gaussian splats, after GSP, are more concentrated towards regions of the image with greater details, allowing {\name} to capture more details.}

\subsection{Dynamic Content\label{sec:dynamic}}

\begin{figure}[t]
    \centering
    \subfloat[Ground Truth]{\parbox{.19\textwidth}{%
        \includegraphics[width=0.18\textwidth]{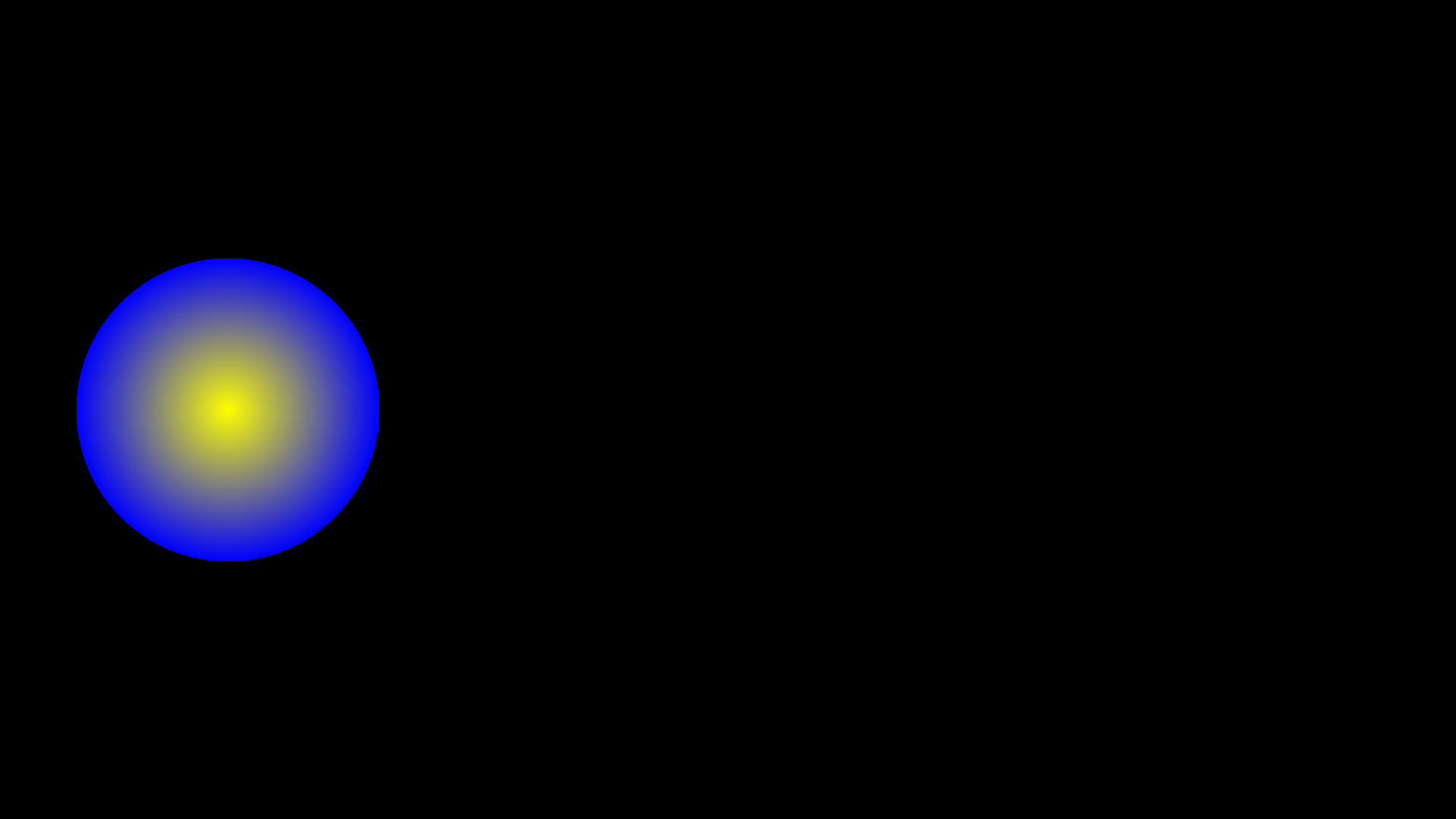}\\
        \includegraphics[width=0.18\textwidth]{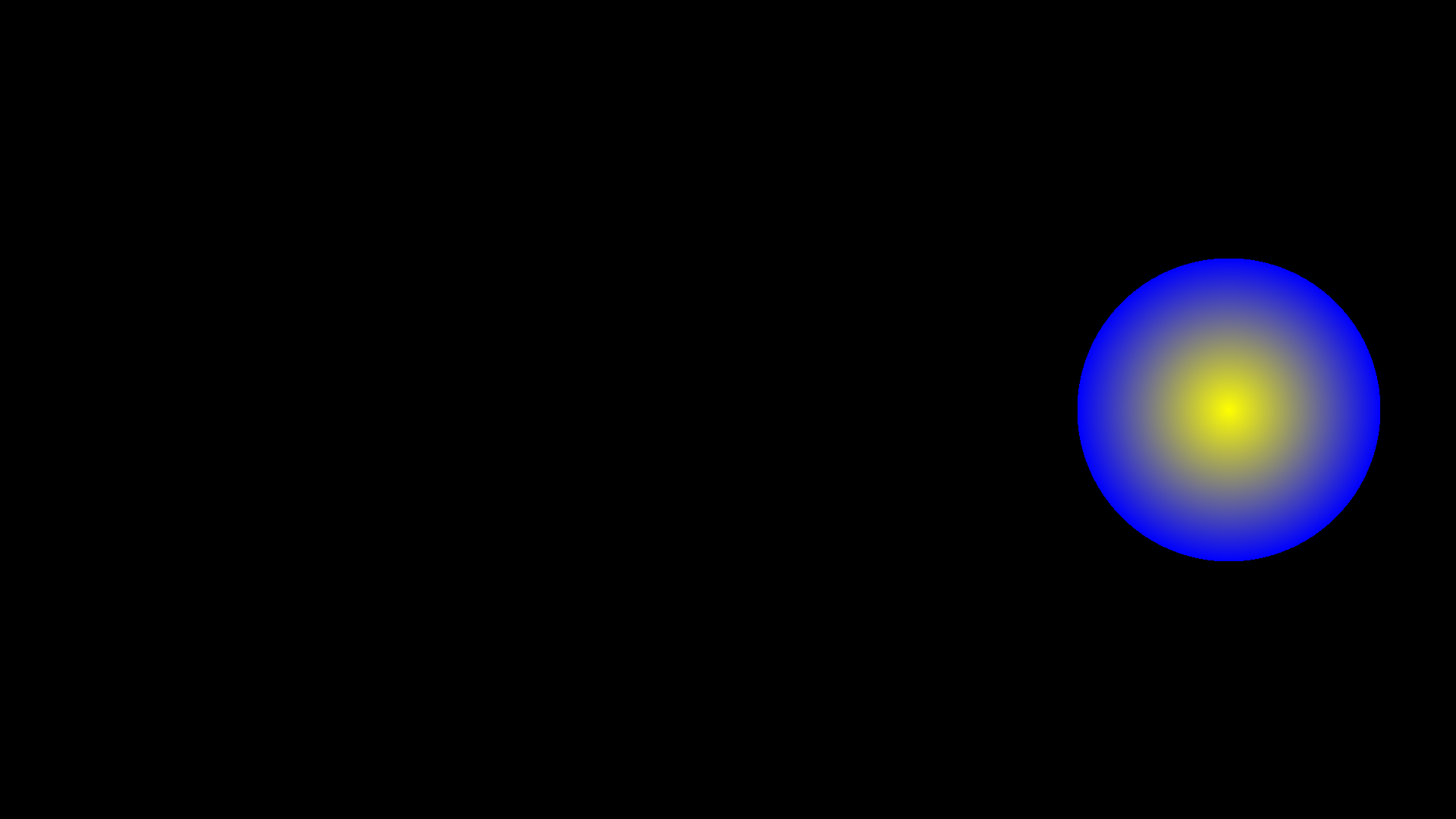}
    }}\hspace{1mm}
    \subfloat[Without GSA]{\parbox{.34\textwidth}{%
        \includegraphics[width=0.36\textwidth]{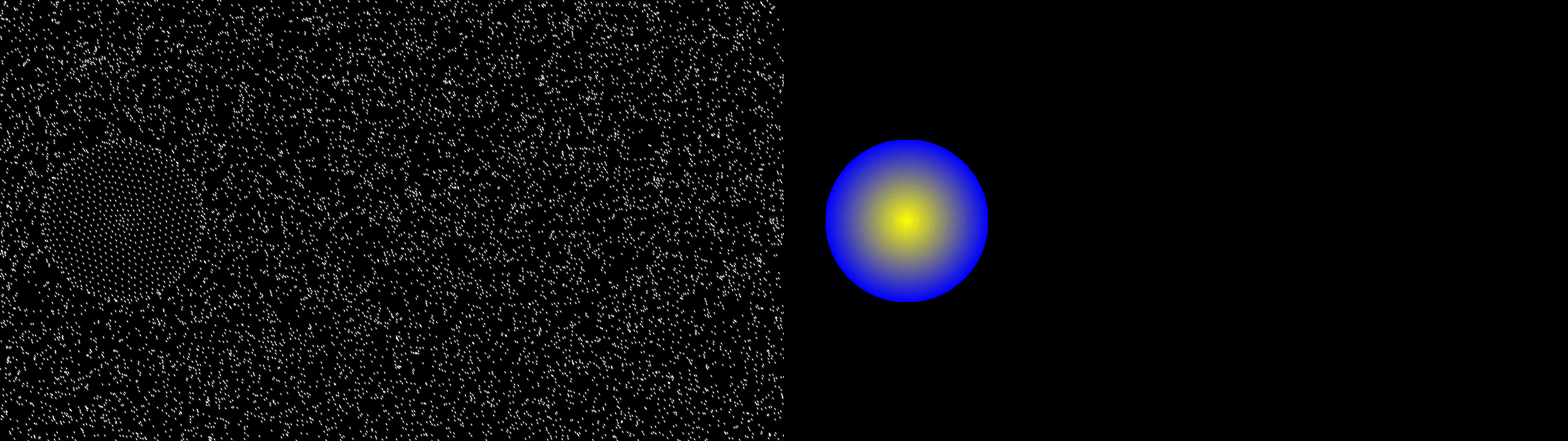}\\
        \includegraphics[width=0.36\textwidth]{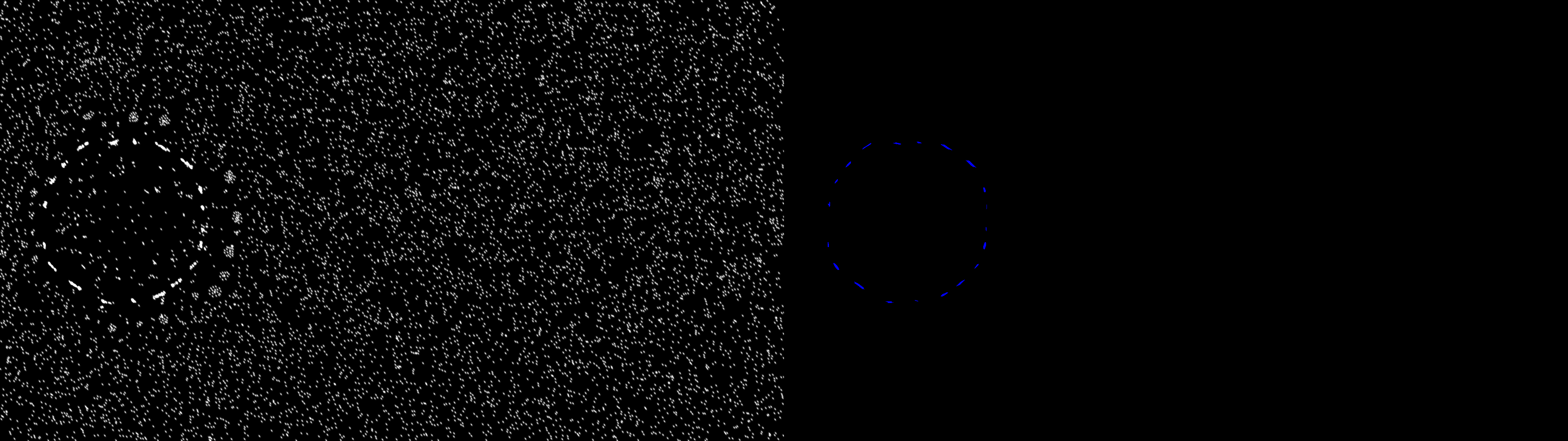}
    }}\hspace{1mm}
    \subfloat[With GSA]{\parbox{.34\textwidth}{%
        \includegraphics[width=0.36\textwidth]{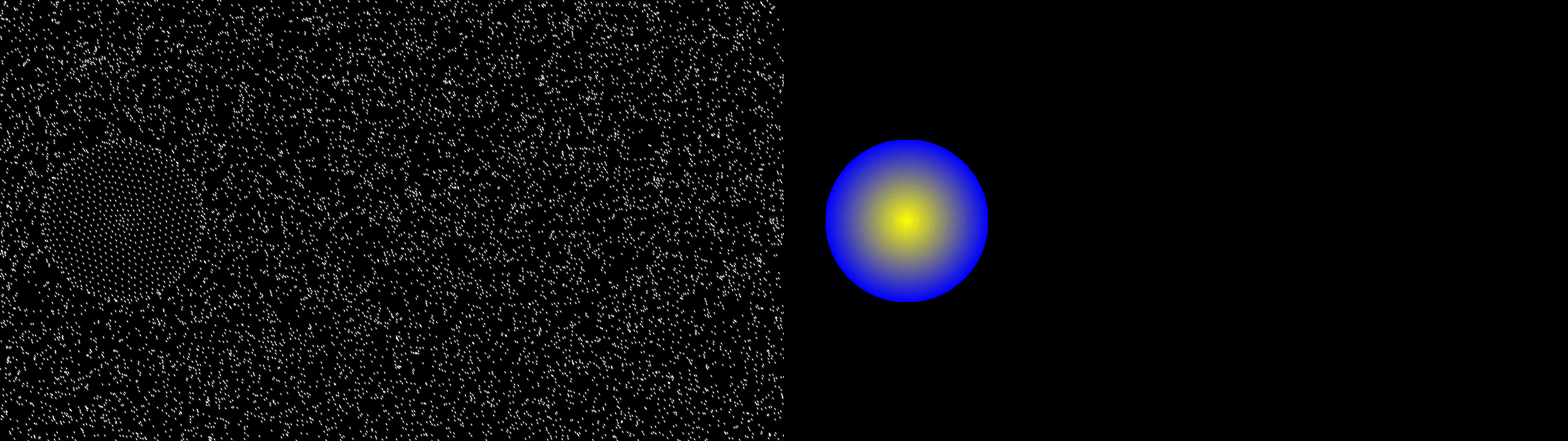}\\
        \includegraphics[width=0.36\textwidth]{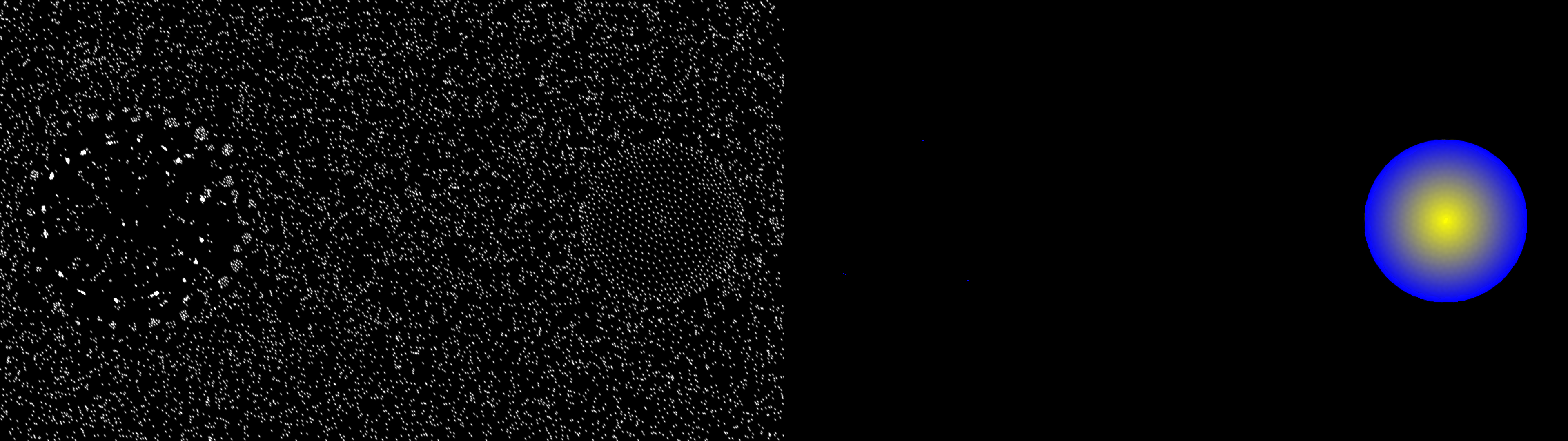}
    }}
    \caption{Distribution of Gaussian Splat Centers on Video Dynamics.}
    \label{fig:Augmenting}
\end{figure}

During training, the Gaussian splats of propagation frames adapt their parameters to changes from the previous frames.  This approach, however, can fail to capture highly dynamic content.  \rev{For example, a new object may appear or existing content may undergo large motion.  Such dynamics may cause the training process to converge to lower frame quality.} 

To address this, we propose Gaussian Splat Augmentation (GSA), which injects new Gaussian splats to better handle dynamics in the frames.
Specifically, in each propagation frame, after initialization using results from the previous frame, every Gaussian splat's contribution $w$ is reset to 1.
New Gaussian splats with $w$ set close to zero are then added uniformly and randomly to the frame.  
During training, we again apply GSP to remove Gaussian splats with low contribution.
This strategy ensures that new Gaussian splats placed over new or dynamic content can be easily activated during training, due to sharp changes in $w$, allowing them to learn and capture the dynamics effectively.
At the same time, splats in well-represented regions with minimal changes in $w$ will be removed by GSP.

Therefore, by coordinating GSA and GSP, our framework can adapt to dynamic content while exploiting temporal redundancy in propagation frames.  These augmented Gaussian splats are analogous to intra-coded macroblocks in P-frames in standard video codecs.  The number of Gaussian splats to augment $N_{aug}$ is a controllable knob to trade off between size and quality, and is dependent on the amount of motion in the scene.

\rev{\Cref{fig:Augmenting} illustrates the impact of the proposed GSA using two frames from a synthetic video (\Cref{fig:Augmenting} (a)), depicting a ball jumping from left to right between consecutive frames.
Figures~\ref{fig:Augmenting} (b) and (c) depict the distributions of Gaussian splat centers and their corresponding rendered images without GSA and with GSA, respectively.  Without GSA, the propagation splats fail to converge to a set of parameters that capture the new position of the circle, resulting in an empty frame.  On the other hand, GSA allows the augmented Gaussian splats to capture the new position of the ball. }

\subsection{Scene Transition Detection \label{sec:keyframe}}
\begin{figure}[t]
    \centering
    \includegraphics[width=0.3\textwidth]{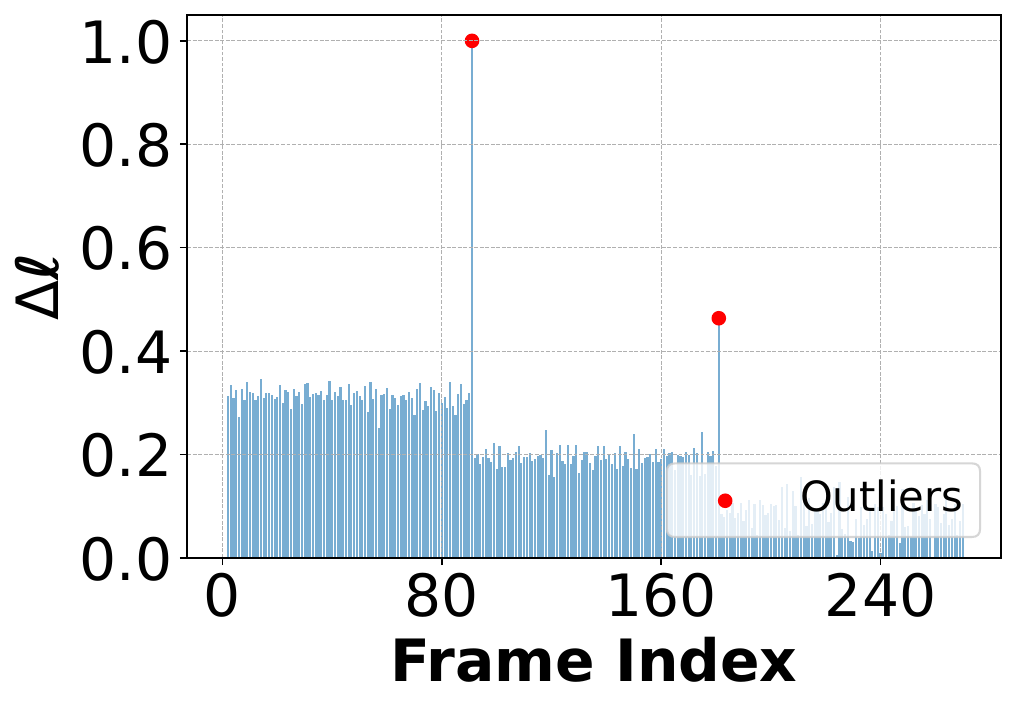}
    \hspace{1mm}
    \includegraphics[width=0.3\textwidth]{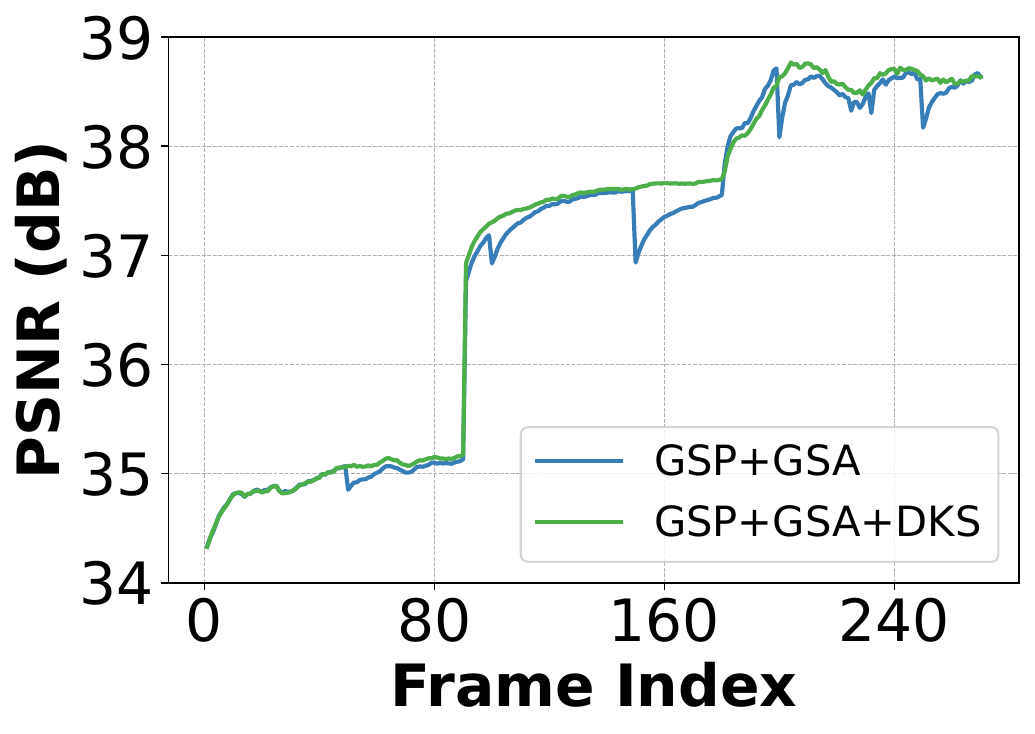}
    \caption{ $\Delta\boldsymbol{\ell}$ and PSNR over Frames on \textit{SceneMix}.}
    \label{fig:DKS}
\end{figure}

While GSA can adapt effectively to frame content dynamics, its effectiveness decreases as the changes across frames become more drastic.  Even if the augmented Gaussian splats can capture the changes, as the differences between the two frames increase, the compression efficiency decreases.  Thus, it is more effective to insert an \textcolor{black}{anchor frame} when there is a significant difference between two adjacent frames, such as when there is a scene change.

We thus propose a Dynamic Key-frame Selector (DKS) method to identify significant frame differences.  
\rev{DKS detects scene transitions from the difference between anchor and propagation frame pre-training losses.
Each frame $f_t$ in the video $\boldsymbol{\mathcal{F}} = \{f_1, f_2, \dots, f_T\}$ is first trained as an \textcolor{black}{anchor frame}, yielding a set of loss values:
    $\boldsymbol{\ell}^{a} = \{\ell^{a}_{1}, \ell^{a}_{2}, \dots, \ell^{a}_{T}\}$.
Subsequently, each frame (except the first) is pre-trained as a \textcolor{black}{propagation frame}, resulting in another set of loss values:
$\boldsymbol{\ell}^{p} = \{\ell^{p}_{2}, \ell^{p}_{3}, \dots, \ell^{p}_{T}\}.$
The loss differences are computed as:}
\begin{equation}
    \label{deltaloss}
   \Delta\boldsymbol{\ell} = \{\Delta\ell_{t} \mid \Delta\ell_{t} = \ell_{t}^{p} - \ell_{t}^{a}\}, \quad t = 2, \dots, T.
\end{equation}

To detect scene transitions, we identify outliers in $\Delta\boldsymbol{\ell}$ by modeling its local distribution using a sliding window. A frame $f_t$ is classified as a scene transition if its loss difference $\Delta\ell_{t}$ exceeds the local mean $\mu_t$ by more than three times the local standard deviation $\sigma_t$ ($\Delta\ell_{t} > \mu_t + 3\sigma_t$), 
where the local mean $\mu_t$ and standard deviation $\sigma_t$ are calculated within a window $\mathcal{W}_t$ centered around $t$:
\begin{equation}
    \mu_t = \frac{1}{|\mathcal{W}_t|} \sum_{i \in \mathcal{W}_t} \Delta\ell_{i}, \quad \sigma_t = \sqrt{\frac{1}{|\mathcal{W}_t|} \sum_{i \in \mathcal{W}_t} (\Delta\ell_{i} - \mu_t)^2}.
\end{equation}
\rev{The window $\mathcal{W}_t$ is defined as:}
\begin{equation}
    \mathcal{W}_t = \{i \mid \max(2, t-w_{\text{win}}) \leq i \leq \min(T, t+w_{\text{win}})\},
\end{equation}
\rev{where $w_{\text{win}}$ denotes the window radius.}
The \textcolor{black}{anchor frame} set $\boldsymbol{\mathcal{A}}$ is constructed by including the first and all frames identified as scene transitions:
\begin{equation}
    \label{Kselect}
    \boldsymbol{\mathcal{A}} = \{1\} \cup \{t \mid \Delta\ell_{t} > \mu_t + 3\sigma_t, \, t = 2, \dots, T\}.
\end{equation}

\Cref{fig:DKS} demonstrates the process of \textcolor{black}{anchor frame} selection.   For demonstration, we synthesize a video by concatenating the first 90 frames of \textit{Beauty}, \textit{HoneyBee}, and \textit{Jockey} from the UVG dataset into one video.  \rev{The figure on the left plots the changes in loss $\Delta \ell_{t}$ across frames.}  We can see two spikes, corresponding to the outliers during scene change (seams when two videos are concatenated).   The plot on the right shows the PSNR of individual frames, comparing when we employ DKS to dynamically decide the \textcolor{black}{anchor frame} versus when we do not use DKS.  In the latter case, we insert an \textcolor{black}{anchor frame} periodically every 50 frames. 
The plot shows that inserting an \textcolor{black}{anchor frame} when there is no scene change lowers frame quality (slight PSNR drop every 50 frames) as {\name} learns the Gaussian splats from scratch.

\begin{algorithm}[t]
\caption{{\name} End-to-End Training Pipeline for 2D Gaussian Video Representation}\label{alg:training}

\begin{algorithmic}[1]
\REQUIRE Video frames $\boldsymbol{\mathcal{F}} = \{f_1, f_2, \dots, f_T\}$, Gaussian splats $\boldsymbol{\mathcal{G}}$;
\\
\STATE Pre-train each frame from scratch to get $\boldsymbol{\mathcal{G}}^{a}$ and compute $\boldsymbol{\ell}^{a}$;
\STATE Pre-train each frame as $\boldsymbol{\mathcal{G}}^{p}$ according to $\boldsymbol{\mathcal{G}}^{a}$ and compute $\boldsymbol{\ell}^{p}$;
\STATE \rev{Select anchor frames to get $\boldsymbol{\mathcal{A}}$ according to~\Cref{Kselect};}
\FOR{$t \in \{1, 2, \dots, T\}$}
    \IF{$f_t \in \boldsymbol{\mathcal{A}}$}
        \STATE Initialize $\boldsymbol{\mathcal{G}}^{*}_t$ with $N+N_{aug}$ random Gaussian splats;
        \WHILE{not converge}
            \STATE Optimize $\boldsymbol{\mathcal{G}}^{*}_t$ according to~\Cref{L2};
        \ENDWHILE
        \STATE \rev{Train the next $k$ frames as propagation frames and obtain $\boldsymbol{\mathcal{G}}^{p}_{t+k}$};
        \STATE \textcolor{black}{Re-initialize} $\boldsymbol{\mathcal{G}}^{*}_t$ according to $\boldsymbol{\mathcal{G}}^{p}_{t+k}$.
    \ELSE
        \STATE Initialize $\boldsymbol{\mathcal{G}}^{*}_t$ using Gaussian splats from $\boldsymbol{\mathcal{G}}^{*}_{t-1}$ and inject $N_{aug}$ random Gaussian splats.
    \ENDIF
    \FOR{$s = 1$ \TO $K_{\mathrm{GSP}}$}
        \STATE Optimize $\boldsymbol{\mathcal{G}}^{*}_t$ according to loss from ~\Cref{L2};
        \IF{$s \bmod Z_{\mathrm{GSP}} = 0$}
            \STATE \rev{Remove $N_{\mathrm{prune}} \cdot \frac{Z_{\mathrm{GSP}}}{K_{\mathrm{GSP}}}$ splats with minimal contribution according to $s_n$ as in~\Cref{eq:importance}};
        \ENDIF
    \ENDFOR
    \WHILE{not converge}
        \STATE Optimize $\boldsymbol{\mathcal{G}}^{*}_t$ according to~\Cref{L2};
    \ENDWHILE
\ENDFOR
\ENSURE Optimized Gaussian splats $\boldsymbol{\mathcal{G}} = \{\boldsymbol{\mathcal{G}}_{1}^{*}, \boldsymbol{\mathcal{G}}_{2}^{*}, \dots, \boldsymbol{\mathcal{G}}_{T}^{*}\}$.
\end{algorithmic}
\end{algorithm}

\subsection{End-to-End Training for 2D Gaussian Video Representation\label{sec:together}}
We now explain how we combine all five mechanisms above into the training process of {\name}.  

Following Zhang \etal~\cite{DBLP:conf/eccv/ZhangGXHWQLGZ24}, we utilize the $L_2$ loss function to minimize the distortion between each original frame $f_t \in \boldsymbol{\mathcal{F}}$ and its reconstructed frame $\widehat{f}_t \in \widehat{\boldsymbol{\mathcal{F}}}$.
The $L_2$ loss function on $f_t$ is given as:
\begin{equation}
\label{L2}
\mathcal{L}_2^{t} = \frac{1}{H \times W} \sum_{i=1}^{H} \sum_{j=1}^{W} \left\| f_t(i, j) - \widehat{f}_t(i, j) \right\|_2^2,
\end{equation}
where $H$ and $W$ are the height and width of the frame (in pixels),  
$f_t(i, j)$ and $\widehat{f}_t(i, j)$ denote the original and reconstructed pixel intensities at position $(i, j)$ in $f_t$ and $\widehat{f}_t$, respectively.  

The training process of {\name} is shown in \Cref{alg:training}.
Lines 1--3 first perform dynamic \textcolor{black}{anchor frame} selection to detect scene change.  We first apply DKS to detect scene transitions: Lines 1 and 2 pre-train each frame under \textcolor{black}{anchor frame} and \textcolor{black}{propagation frame} configurations to obtain their reconstruction losses, and Line 3 selects \textcolor{black}{anchor frames} according to the loss difference.
\rev{For each frame $f_t$, if $f_t$ is an \textcolor{black}{anchor frame}, we train the Gaussian splats for the frame, starting from random Gaussian splats (Line 6), and optimize them until convergence (Lines 7--9).  This line corresponds to the first pass.}
\textcolor{black}{We then perform the second pass by predicting the next $k$ frames (Line 10), re-initializing the \textcolor{black}{anchor frame} with the $k$-th frame (Line 11).  The algorithm then re-optimizes the \textcolor{black}{anchor frame} in Lines 15--23 with GSP to remove splats with minimal contribution.}

\rev{If $f_t$ is a \textcolor{black}{propagation frame}, then the algorithm performs frame prediction from its previous reconstructed frame and adds new Gaussian splats for dynamics through GSA (Line 13).}
Then, the algorithm applies GSP during the first $K_{\mathrm{GSP}}$ iterations of training.  \rev{The algorithm removes $N_{prune}$ Gaussian splats during this process.}  This removal is performed periodically, after every $Z_{\mathrm{GSP}}$ iterations, as shown in Lines 17--19.  After GSP, the training process continues until the Gaussian splats converge (Lines 21--23).  

To keep the number of Gaussian splats in each frame constant as $N$, we can set $N_{prune} = N_{aug}$.

\section{Video Compression through 2D Gaussian Splats}
In this section, we introduce how {\name} compresses the 2D Gaussian video representation into a compact bitstream by quantization. 
Drawing on standard video coding terminology, we adopt the term intra-coded frame (I-frame) and predictive-coded frame (P-frame).  Each anchor frame is encoded as an I-frame, where we store the absolute Gaussian splat parameters. 
To improve error robustness and limit the propagation of quantization errors, we periodically intra-code a propagation frame as an I-frame. The remaining propagation frames are encoded as P-frames, in which only the changes of the propagation splats are stored.
We first describe the I-frame quantization strategy in~\Cref{sec:I-frame_quantization}, then present the residual-based P-frame quantization strategy in~\Cref{sec:P-frame_quantization}.

\subsection{I-frame quantization \label{sec:I-frame_quantization}}
\begin{figure}[t]
    \centering
    \begin{minipage}[b]{\textwidth}\centering
        \subfloat[Encode]{\includegraphics[width=0.4\textwidth]{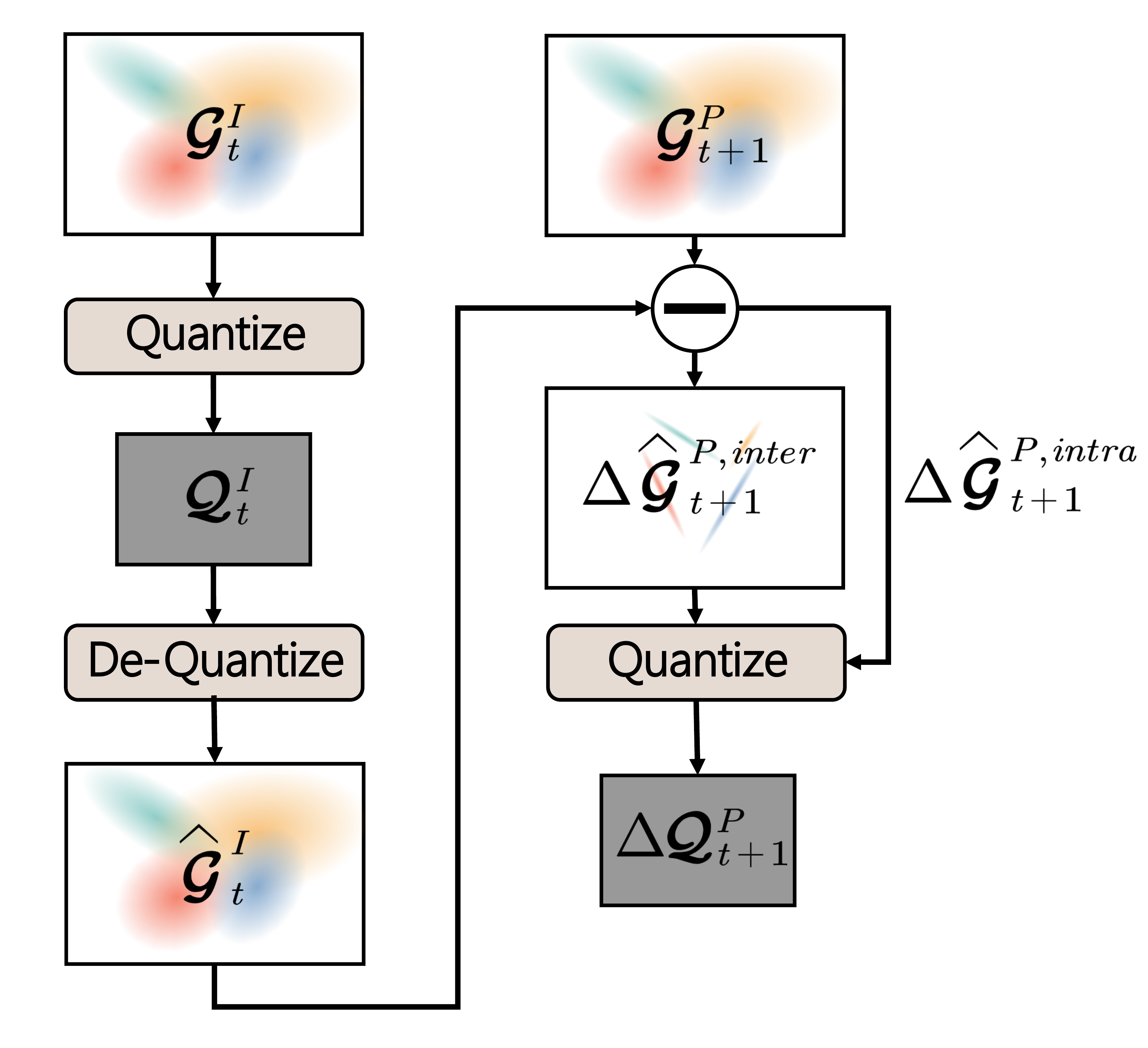}\label{fig:encode}}
        \hspace{0.05\textwidth}
        \subfloat[Decode]{\includegraphics[width=0.4\textwidth]{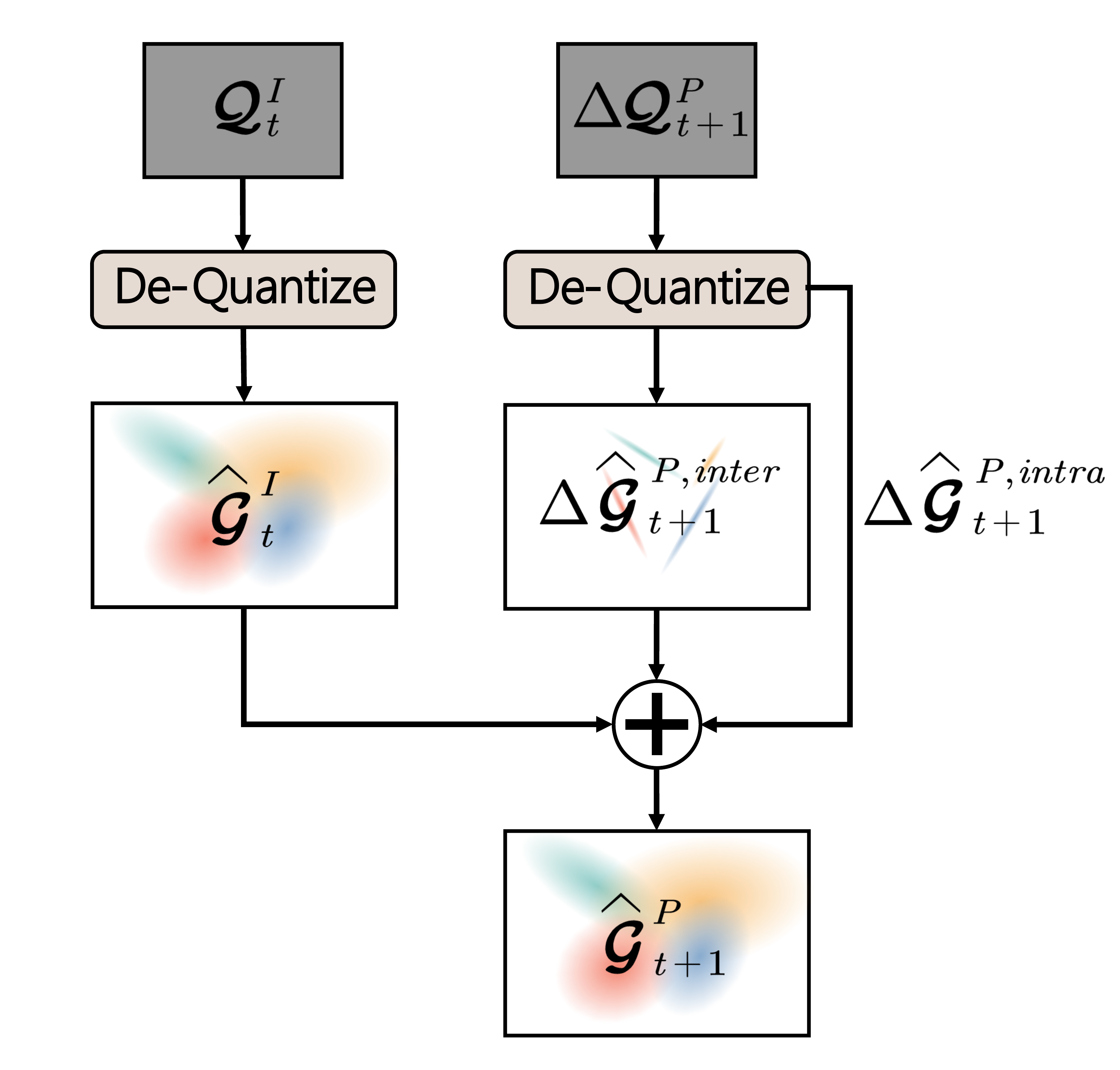}\label{fig:decode}}
    \end{minipage}
    \caption{Quantization Pipeline. (a) shows the encoding pipeline, and (b) shows the decoding pipeline. We store the quantized I-frame Gaussian splats, $\boldsymbol{Q}^{I}$, and the quantized differences of the P-frame Gaussian splats, $\Delta\boldsymbol{Q}^{P}$.}
    \label{fig:quantization}
\end{figure}

As shown in~\Cref{fig:encode}, once the video has been represented as a set of Gaussian splats $\boldsymbol{\mathcal{G}}$, we apply a temporal-aware fine-tuning strategy and then quantize the Gaussians. For I-frames, we quantize the full set of Gaussian anchor splats, obtaining $\boldsymbol{\mathcal{Q}}^{I}$.

For the $i$-th frame, if it is an I-frame, its $n$-th Gaussian splat $\boldsymbol{\mathcal{G}}^I_{i,n} = \{\boldsymbol{\mu}_n, \boldsymbol{\ell}_n, \boldsymbol{c}'_n\}$ is quantized as:
\begin{equation}
\boldsymbol{\mathcal{Q}}^{I}_{i,n} = \mathcal{Q}^I(\boldsymbol{\mathcal{G}}^{I}_{i,n}),
\end{equation}
\rev{where $\mathcal{Q}^I(\cdot) = \{\mathcal{Q}^I_{\mu}(\cdot), \mathcal{Q}^I_{\ell}(\cdot), \mathcal{Q}^I_{c}(\cdot)\}$ denotes the quantization operator based on the Attribute Quantization-aware Fine-tuning~\cite{DBLP:conf/eccv/ZhangGXHWQLGZ24} designed for 2D Gaussian splats.}
\rev{Specifically, for position $\boldsymbol{\mu}$, 16-bit float precision is used as $\mathcal{Q}_{\mu}(\boldsymbol{\mu})$.}
For the Cholesky vector, $\boldsymbol{\ell}$, $b$-bit asymmetric quantization~\cite{DBLP:journals/corr/abs-2004-09576} is applied:  
\begin{equation}
\begin{array}{c} 
    \mathcal{Q}^I_{\ell}(\ell_i) = \hat{\ell}_i \times \gamma_i + \beta_i \\
    \hat{\ell}_i = \left\lfloor \text{clamp}\left(\frac{\ell_i - \beta_i}{\gamma_i}, 0, 2^b - 1\right) \right\rfloor,
\end{array}
\end{equation}
where $i \in \{0, 1, 2\}$, and $\gamma_i, \beta_i$ are learnable scaling and offset factors.
During fine-tuning, we learn the scaling and offset factors and quantize the covariance parameters to $b$ bits, while the corresponding scaling and offset factors for de-quantization are kept in 32-bit floating-point format.

\rev{For the weighted color $\boldsymbol{c}'$, we apply $M$-stage residual vector quantization (RVQ)~\cite{DBLP:journals/taslp/ZeghidourLOST22}, where each stage uses a VQ codebook~\cite{gray1984vector} of size $B$:}
\begin{equation}
\begin{array}{c}
\mathcal{Q}^I_{c}(\boldsymbol{c}')
=
\sum_{k=1}^{M}\boldsymbol{C}^{k}[i^{k}],\\[2mm]
i^{k}
=
\arg\min_{j\in\{1,\ldots,B\}}
\left\|
\boldsymbol{C}^{k}[j]
-
\left(
\boldsymbol{c}'
-
\sum_{r=1}^{k-1}\boldsymbol{C}^{r}[i^{r}]
\right)
\right\|_{2}^{2},
\end{array}
\end{equation}
\rev{where $\boldsymbol{C}^{k}\in\mathbb{R}^{B\times 3}$ denotes the codebook at stage $k$, and $i^{k}$ is the selected codeword index. Each stage quantizes the residual left by the preceding stages, and the final color is reconstructed by summing the selected codewords. The codebooks are jointly optimized using the commitment loss~\cite{DBLP:conf/eccv/ZhangGXHWQLGZ24}.
Note that Gaussian colors are highly compressible.
For example, in I-frames, two codebooks of size 8 require only 16 learned color vectors and provide 64 possible color combinations. However, when Gaussians overlap, their color contributions accumulate, allowing these few basis colors to synthesize a wide range of rendered colors.
Meanwhile, when decoding an I-frame, the decoder directly dequantizes its bitstream $\boldsymbol{\mathcal{Q}}^{I}$ to reconstruct the full set of Gaussian splats $\hat{\boldsymbol{\mathcal{G}}}^{I}$ as shown in~\Cref{fig:decode}.}

\subsection{P-frame quantization \label{sec:P-frame_quantization}}
For each P-frame, since it is initialized by inheriting Gaussian splats from its previous frames, we can track the Gaussian splats' deltas, referring to its previous frame. 
\rev{Since each Gaussian is tied to a local surface patch and encodes both geometry (position, scale, and orientation) and color, a Gaussian splat that fits a region in one frame usually remains a good approximation for nearby frames in natural videos, requiring only small parameter adjustments to account for motion and illumination changes.}
These per-Gaussian temporal residuals are thus inherently low-magnitude and highly compressible, so we quantize them to obtain $\Delta\boldsymbol{\mathcal{Q}}^{P}$.
The bitstream therefore stores the quantized I-frame representation $\boldsymbol{\mathcal{Q}}^{I}$ together with the quantized P-frame residuals 
$\Delta\boldsymbol{\mathcal{Q}}^{P}$.

In P-frames, by tracking the deltas of Gaussian splats across time, we can distinguish two types of Gaussian updates:
(i) \rev{existing Gaussian parameter deltas, or the deltas of propagation splats, $\Delta\boldsymbol{\mathcal{G}}^{P,\text{inter}}$, which describe parameter differences with respect to their corresponding Gaussians in the previous frame}; and (ii) changes caused by GSA and GSP, $\Delta\boldsymbol{\mathcal{G}}^{P,\text{intra}}$, which encode the removal and insertion of Gaussian splats within the current frame.
As these two components exhibit distinct statistical distributions, we quantize them separately, which enables more precise bit allocation across different temporal changes and leads to better compression efficiency.
The overall encoding pipeline is illustrated in~\Cref{fig:encode}.
Formally, for the $i$-th P-frame, we define:
\begin{equation}
\begin{gathered}
\Delta\boldsymbol{\mathcal{Q}}^{P}_{i}
= \mathcal{Q}^{P,\text{inter}}_{i} \cup \mathcal{Q}^{P,\text{intra}}_{i}, \\[4pt]
\mathcal{Q}^{P,\text{inter}}_{i}
= \Big\{
      \mathcal{Q}^P\big(\Delta\boldsymbol{\mathcal{G}}^{P,\text{inter}}_{i,n}\big)
      \;\Big|\;
      \Delta\boldsymbol{\mathcal{G}}^{P,\text{inter}}_{i,n}
      = \boldsymbol{\mathcal{G}}^{P}_{i,n} - \boldsymbol{\hat{\mathcal{G}}_{i-1,n}},\;
      n \in \mathcal{N}^{\text{inter}}
   \Big\},\\[4pt]
\mathcal{Q}^{P,\text{intra}}_{i}
= \Big\{
    \mathcal{Q}^I\big(\Delta\boldsymbol{\mathcal{G}}^{P,\text{intra}}_{i,n}\big)
      \;\Big|\;
      \Delta\boldsymbol{\mathcal{G}}^{P,\text{intra}}_{i,n}
      = \boldsymbol{\mathcal{G}}^{P}_{i,n},\;
      n \in \mathcal{N}^{\text{intra}}
   \Big\}.
\end{gathered}
\end{equation}
\textcolor{black}{Here,  $\mathcal{N}^{\text{inter}}$ is the index set of inter-coded splats, and $\mathcal{N}^{\text{intra}}$ is the index set of anchor splats, or intra-coded splats.}
For \textcolor{black}{$\Delta\boldsymbol{\mathcal{G}}^{P,\text{inter}}$}, each element represents only parameter differences, which are typically smaller in magnitude and therefore allow more compact encoding.
In practice, we quantize the position deltas with 12-bit precision (instead of the 16-bit precision used for absolute positions in I-frames), allocate fewer bits to the Cholesky-vector deltas of the covariance, and use a smaller codebook for the color residuals.
For the Cholesky-vector deltas, we quantize into the signed integer range $\left[-2^{b-1}, 2^{b-1}-1\right]$, since these deltas can be both positive and negative and are often centered around zero, making a symmetric signed range more effective than an unsigned one.
For \textcolor{black}{$\Delta\boldsymbol{\mathcal{G}}^{P,\text{intra}}$}, since each element corresponds to a complete Gaussian splat, it is quantized in the same manner as $\boldsymbol{\mathcal{G}}^{I}$.

\rev{As shown in~\Cref{fig:decode}, to decode the P-frame, the decoder first reconstructs the reference frame by de-quantizing its previous frames.
Then, we de-quantize  $\Delta\boldsymbol{\mathcal{Q}}^{P}_{t+1'}$ to recover the parameter deltas \textcolor{black}{$\Delta\hat{\boldsymbol{\mathcal{G}}}^{P,\text{inter}}_{t+1'}$ of inter-coded splats} and the \textcolor{black}{newly added intra-coded splats} $\Delta\hat{\boldsymbol{\mathcal{G}}}^{P,\text{intra}}_{t+1'}$.
The existing deltas are added back to their corresponding Gaussian splats in the previous decoded frame, whereas the new Gaussian splats directly replace those in the reference frame that are about to be removed.
The union of these two parts forms the reconstructed P-frame $\hat{\boldsymbol{\mathcal{G}}}^{P}_{t+1'}$.}

\subsection{Stored Parameters for I-frames and P-frames}

\begin{table}[t]
\centering
\caption{\rev{Stored parameters and side information for I-frames and P-frames.}}
\label{tab:stored_params}
\begin{tabular}{lccc}
\toprule
\textbf{Frame Type} & \textbf{I-frame} & \textbf{P-frame} \\
\midrule
Covariance scaling/offset & 24 bytes& 48 bytes \\
Color codebooks & 192 bytes& 360 bytes \\
\textbf{Total Per-frame Info} & \textbf{216 bytes}  & \textbf{408 bytes} \\

\midrule
\textbf{Splat Type} & \textbf{\rev{Intra-coded Splats (Anchor/Propagation)}} & \textbf{\rev{Inter-coded Splats (Propagation)}} \\
\midrule
Position &
\rev{32 bits} &
\rev{24 bits} \\

Quantized Cholesky vector  &
\rev{18 bits} &
\rev{15 bits} \\

Color Index &
6 bits &
6 bits \\

\textbf{Total Per-splat Size} &
\rev{\textbf{56 bits}} &
\rev{\textbf{45 bits}} \\
\bottomrule
\end{tabular}
\end{table}

To better illustrate the structure of the compressed bitstream, \Cref{tab:stored_params} summarizes all stored parameters for I-frames and P-frames under our residual-based quantization framework.

\rev{For an I-frame, all Gaussians are encoded as intra-coded splats. To enable de-quantization, we store the scaling and offset factors for decoding the covariance matrix and codebook for decoding color.}
Specifically, each factor is represented in 32-bit floating-point format. Since the Cholesky vector of the covariance matrix has three components, we store three pairs of scaling and offset values, resulting in $3 \times (32 + 32) = 192~\text{bits} = 24~\text{bytes}$.
Two color codebooks are used for the three RGB channels. Each codebook contains 8 entries stored in 32-bit floating-point format, so the total storage is $3 \times 2 \times 8 \times 32 = 1536~\text{bits} = 192~\text{bytes}$.
Each intra-splat stores its position using 16 bits, the quantized Cholesky vector using 6 bits, and the color index using 6 bits, giving a per-splat total of \rev{$16\times2 + 6\times3 + 6 = 56~\text{bits}$.}

\rev{For a P-frame, we store both inter-coded splats and intra-coded splats.} 
A P-frame also requires decoding parameters for both splat types. 
\rev{For inter-coded splats, we maintain an additional color codebook with 7 entries, whose storage cost is $2 \times 7 \times 3 \times 32 = 1344~\text{ bits } = 168~\text{bytes}$.}
We also store another set of scaling and offset factors (three 32-bit pairs), identical to the I-frame case.  
\rev{For intra-coded splat decoding, the storage matches that of the I-frame (192~bytes).  
Thus, the total P-frame side information amounts to
$24~\text{bytes} + 168~\text{bytes} + 216~\text{bytes} = 408~\text{bytes}$.}
Inter-splats are encoded more compactly because only residual updates are stored. 
We use 12 bits for position deltas and 5 bits for the quantized Cholesky vector, giving \rev{$12\times2 + 5\times3 + 6 = 45~\text{bits}$}.
\rev{In contrast, intra-coded splats in P-frames use the same representation as I-frame intra-coded splats, which is 56 bits per splat.}

\rev{Finally, when forming the final bitstream, we further apply entropy coding to the quantized Cholesky vectors and color indices, further reducing the compressed size.}
\section{Experiments}

\subsection{Experimental Setup}
\rev{We evaluate on two widely used video coding datasets: UVG~\cite{DBLP:conf/mmsys/MercatVV20} and HEVC Class B~\cite{bossen2010common}.
All evaluated sequences have a resolution of $1920\times1080$, 4:2:0 chroma subsampling, and 8-bit depth in YUV format.
We use the first 50 frames of each UVG sequence and the first 96 frames of each HEVC Class B sequence.
The two datasets provide complementary evaluation conditions: UVG (120 fps) allows us to assess the exploitation of temporal coherence, while HEVC Class B (24--60 fps) introduces potentially larger inter-frame displacements, providing a more challenging setting for Gaussian propagation.}

\textbf{Metrics}.
We evaluate our method using three widely adopted metrics: Peak Signal-to-Noise Ratio (PSNR), Multi-Scale Structural Similarity Index Measure (MS-SSIM)~\cite{wang2003multiscale}, and Video Multi-Method Assessment Fusion (VMAF)~\cite{vmaf}.
PSNR quantifies reconstruction fidelity in decibels (dB) by comparing the maximum signal power to noise.
MS-SSIM measures perceptual similarity based on luminance, contrast, and structure across multiple scales, with values ranging from 0 to 1, where higher indicates better similarity.
VMAF combines human visual perception models to provide a comprehensive assessment of video quality, offering a score from 0 to 100, where higher scores denote better quality.

\textbf{Implementation Details}.
\rev{Our {\name} framework is built on the gsplat library~\cite{DBLP:journals/corr/abs-2409-06765} and utilizes CUDA-based rasterization optimized with accumulated summation~\cite{DBLP:conf/eccv/ZhangGXHWQLGZ24}.
Gaussian parameters are optimized using the Adan optimizer~\cite{DBLP:journals/pami/XieZLLY24} with an initial learning rate of $1e^{-3}$, halved every 20,000 steps.
Pre-training is performed with 1,000 iterations for anchor frames and 10 for propagation frames.
GSP prunes 10\% of the $N$ splats in the anchor frames after training.
To eliminate dependence on random seeds, we use unscrambled Sobol sequences~\cite{sobol1967distribution} for all uniform sampling during Gaussian initialization.
For intra-coded Gaussians, weighted colors are initialized to zero, while Cholesky parameters are sampled with $\ell_1 \in [0,1]$ and $\ell_0,\ell_2 \in [0.5,1]$ to avoid overly small or elongated Gaussians and improve initial coverage.
For anchor frames, $K_{GSP}$ is set to 4,000, and for propagation frames, it is set to 1,000. For both frame types, $Z_{GSP}$ is set to 100.
For propagation frames, GSA adds 10\% new splats and GSP subsequently prunes 10\%, maintaining the target Gaussian budget.}
We train the models for 50,000 iterations, or until the model has converged.  The convergence is defined as 100 iterations with at most $1e^{-9}$ changes in the $L_2$ loss between two iterations.
\rev{We use constant-rate coding for our experiments.  To understand the performance of {\name} over different bit rates, we vary $N$ from 10,000 to 50,000, in increments of 10,000.}
The window size $w_{\text{win}}$ is set to 10.
For quantization, we use the same configuration for I-frames and anchor splats in P-frames, \textcolor{black}{which are all intra-coded splats}: the covariance parameters are quantized with precision $b = 6$ bits, and the RVQ color vectors are quantized with a codebook size $B = 8$ and $M = 2$ residual quantization stages.
\rev{For inter-coded splats in P-frames, whose parameters are updated over time, we adopt a more compact setting by quantizing only their parameter deltas: the covariance \emph{updates} are quantized with $b = 5$ bits, and the RVQ color \emph{residuals} are quantized with $(B, M) = (7, 2)$.}
\textcolor{black}{During quantization}, for \textit{Beauty}, \textit{HoneyBee}, \textit{Bosphorus}, \textit{ReadySetGo}, and \textit{ShakeNDry} we place an \textcolor{black}{additional} I-frame every 4 P-frames, whereas for \textit{Jockey} and \textit{YachtRide} we insert them more frequently, using one I-frame for every 2 P-frames, as they contain faster motion with less temporal redundancy.
During storage, we save the learned scale and offset parameters, the RVQ codebooks, the quantized coordinates, and the covariance parameters, as well as the color codebook indices.
Following GaussianImage~\cite{DBLP:conf/eccv/ZhangGXHWQLGZ24}, the bitstream size of the video is computed after applying entropy coding to the quantized covariance parameters and the color indices.
\rev{To reduce CPU launch overhead, we employ CUDA Graphs~\cite{nvidia_cuda_graphs} to capture and replay repeated optimization steps.
Graph replay begins once the Gaussian counts are fixed during representation optimization for each frame, and after initialization during quantization-aware fine-tuning.}
All experiments are carried out on NVIDIA H100 GPUs.

\textbf{Baseline}.  We compare against three groups of existing methods.  First, we benchmark against widely used implementations of standard video codecs, including H.264/AVC~\cite{DBLP:journals/tcsv/WiegandSBL03}, H.265/HEVC~\cite{DBLP:journals/tcsv/SullivanOHW12}, H.266/VVC~\cite{DBLP:conf/vcip/Sullivan20}, and AV1~\cite{chen2020overview}.  
\rev{For AVC and AV1, we use the \texttt{ffmpeg} implementation (v5.1), varying the CRF (\texttt{--crf}) parameter from 30 to 10 to assess compression quality. }
For HEVC and VVC evaluations, we employ HM-16.2 and VTM-23.6, respectively. Both operate in random-access mode, with the QP (\texttt{--QP}) parameter varied between 28 and 18.
Default encoding parameters are used unless otherwise specified to ensure consistency across evaluations. 
Second, we compare our methods against two recent neural video compression methods, DCVC-HEM~\cite{DBLP:conf/mm/Li0022} and DCVC-DC~\cite{DBLP:conf/cvpr/LiLL23}. 
\rev{Finally, we use GaussianImage~\cite{DBLP:conf/eccv/ZhangGXHWQLGZ24} to encode each frame independently as a baseline for a naive Gaussian-splat-based video representation. }
In this baseline, all frames are anchor frames, encoded as I-frames.  Each frame is reconstructed independently, without frame prediction, two-pass encoding, GSP, GSA, or DKS.  As a result, all Gaussian splats are quantized in the same way.
We denote this baseline as GI in the experiments below. 
\rev{For a fair comparison, all applicable methods use the same GOP structure as {\name}.}
We did not compare against the 3DGS-based methods, VeGaS~\cite{DBLP:journals/corr/abs-2411-11024} and VGR~\cite{DBLP:conf/nips/Sun0MLC024}, as they are designed to support video editing rather than compression, and thus their sizes are not comparable.

\subsection{Results and Evaluation \label{sec:results_evaluation}}

\begin{figure}[t]
    \centering
    \subfloat[UVG\label{fig:rdcurve_uvg}]{%
        \parbox{\textwidth}{%
            \centering
            \includegraphics[width=0.25\textwidth]{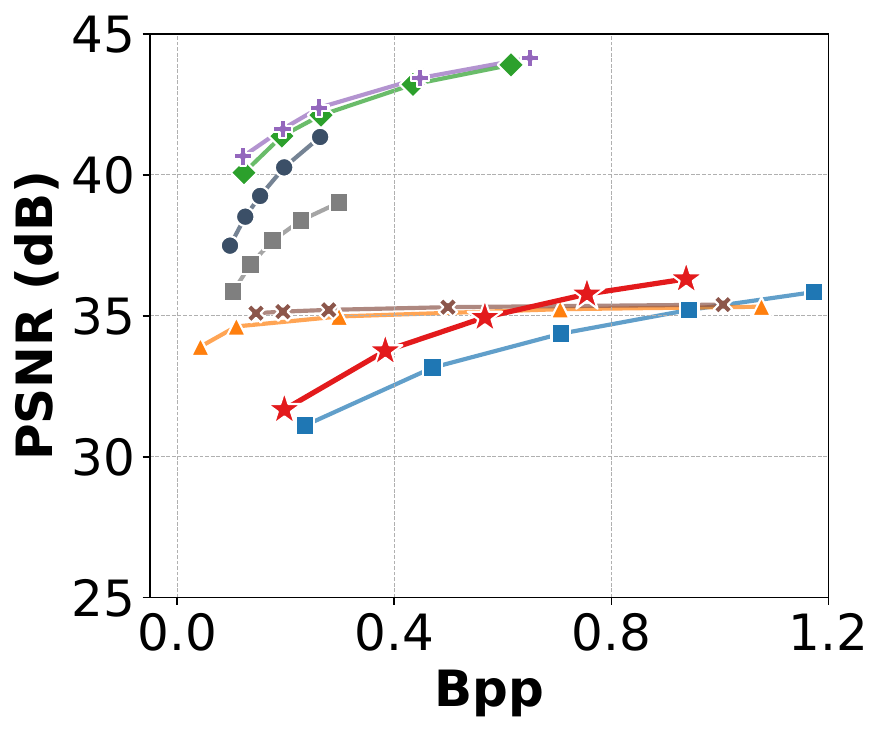}
            \includegraphics[width=0.25\textwidth]{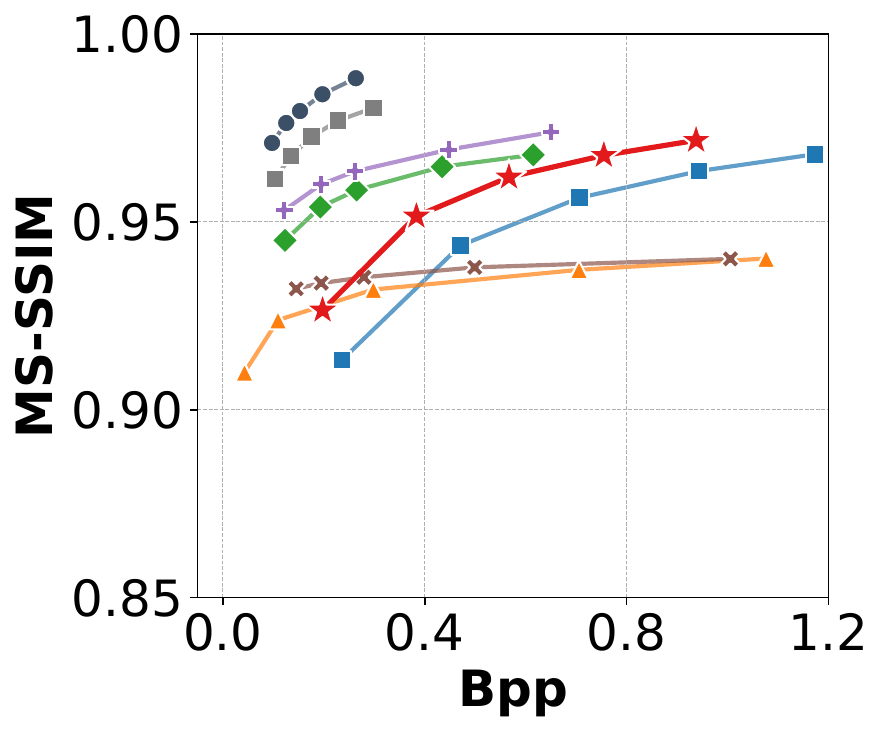}
            \includegraphics[width=0.335\textwidth]{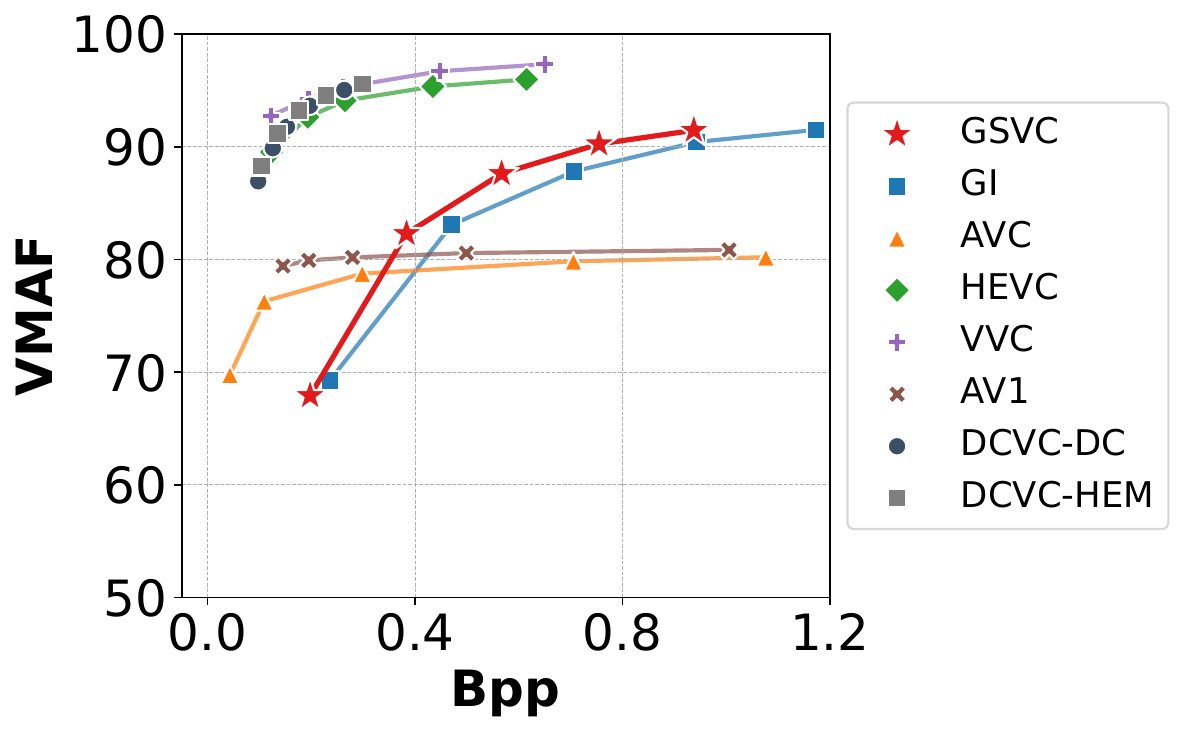}
        }
    }
    \hfill
    \subfloat[HEVC Class B\label{fig:rdcurve_hevc}]{%
        \parbox{\textwidth}{%
            \centering
            \includegraphics[width=0.25\textwidth]{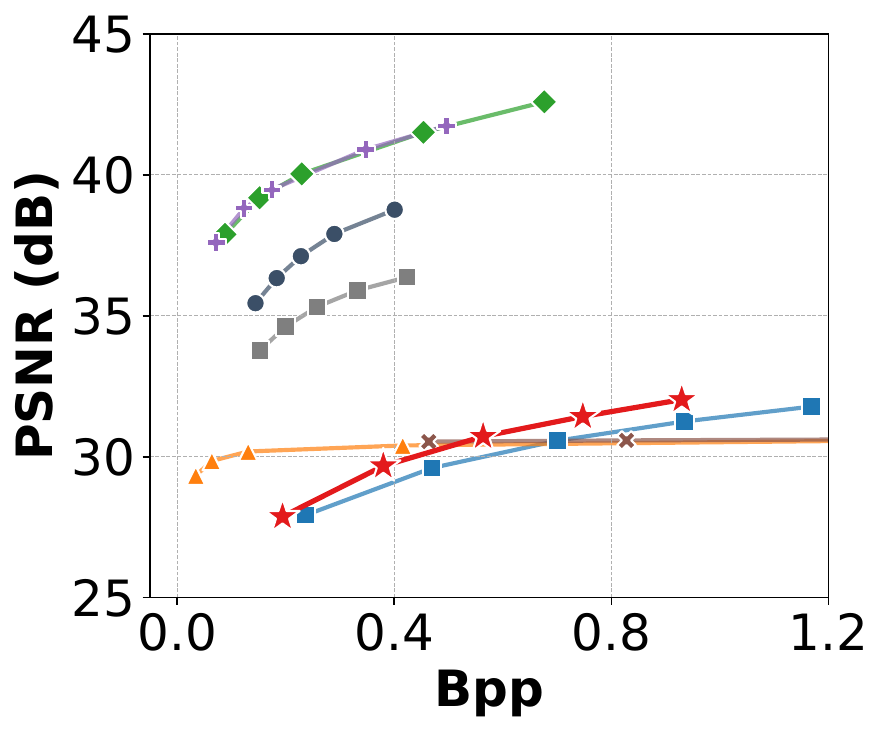}
            \includegraphics[width=0.25\textwidth]{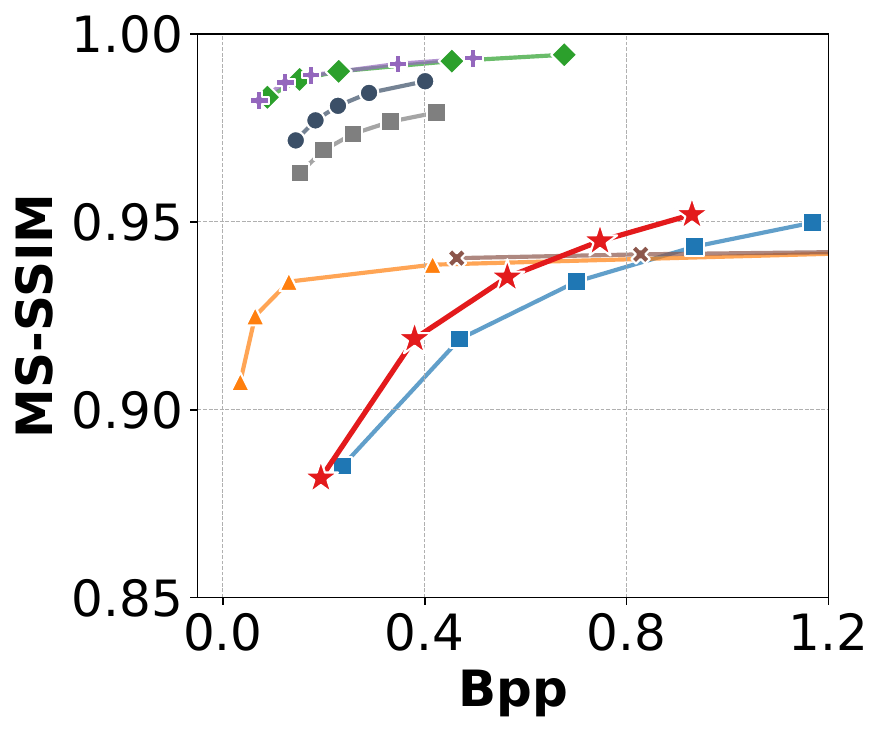}
            \includegraphics[width=0.335\textwidth]{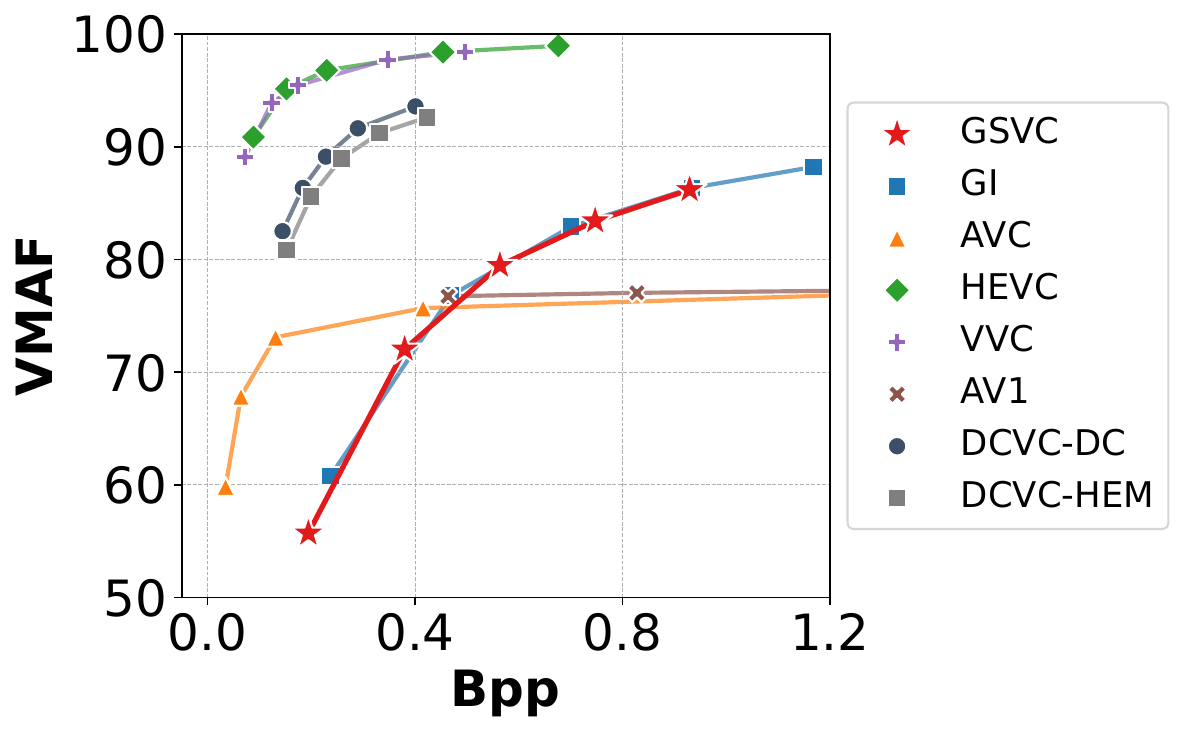}
        }
    }

    \caption{\rev{Rate-distortion performance of GSVC and the baselines in terms of PSNR, MS-SSIM, and VMAF.}}
    \label{fig:rdcurve}
\end{figure}

\rev{\textbf{Rate-Distortion Trade-off}.
\Cref{fig:rdcurve} presents the key results, showing the rate-distortion curves of {\name} and the baselines.
{\name} substantially outperforms the GaussianImage-based baseline on UVG, demonstrating the benefit of exploiting temporal redundancy through Gaussian propagation and residual-based quantization. On HEVC Class B, {\name} also achieves better PSNR and MS-SSIM with comparable VMAF, showing that these gains extend to sequences with more complex content and larger inter-frame motion.}

\rev{The neural compression methods achieve higher MS-SSIM than {\name} and, at some operating points, standard codecs such as VVC.
These methods rely on pre-trained models and control rate-quality through learned quantization parameters.
In contrast, like standard MPEG/H.26x codecs, {\name} requires no pre-training and can be directly applied to unseen videos. Moreover, {\name} supports arbitrary Gaussian budgets, providing fine-grained control over the rate-distortion trade-off.}

\rev{Compared with standard video codecs, {\name} achieves better than or comparable performance in terms of MS-SSIM and VMAF (which consider image structure and human perception) to AVC and AV1, while attaining MS-SSIM performance comparable to HEVC and VVC.
For HEVC and VVC, we use the reference encoders HM-16.2 and VTM-23.6, respectively, with well-tuned configurations. In contrast, AVC and AV1 are evaluated using their \texttt{ffmpeg} implementations, which are optimized for speed instead of rate-distortion performance, to keep our evaluation setup aligned with common usage.
Our experiments on {\name} yield a lower PSNR in some scenarios.  However, it is important to note that we solely use $L_2$ loss as the loss function for optimization, and we can tune the convergence condition to further improve the PSNR, without increasing $N$, if we sacrifice the encoding time.}

\rev{\textbf{Computational Cost}.
Under the Gaussian budgets evaluated in our experiments, {\name} requires an average encoding time of 5--7 minutes per frame across the UVG and HEVC Class B datasets, including both representation optimization and quantization.
At a playback rate of 24 fps, its bitrates of 0.19--0.94 bpp correspond to data rates of 1.18--5.85 MB/s. 
{\name} achieves rendering speeds of 1,413--1,539 fps under these Gaussian budgets. }

\rev{Considering decoding efficiency, we first compare {\name} with VVC on the same Intel Core Ultra 5 125H laptop CPU under Windows.
{\name} requires 39.89--65.09 ms per frame, compared with 60.75--141.79 ms for VVC.
Using Windows E3 application-attributed CPU energy estimates, {\name} consumes 1.59--2.13 J per frame with eight threads, compared with 0.51--1.31 J per frame for VVC at QPs 18 and 28.
We further optimize the CPU decoder for {\name} on an Apple M3 Pro running macOS, achieving a decoding latency of 2.9--5.7 ms per frame and per-process CPU energy consumption of 64.0--102.8 mJ per frame.
The per-process CPU energy estimate on macOS is obtained using the \texttt{proc\_pid\_rusage} API and reported as the median over three runs.
Using this optimized decoder, {\name} is estimated to consume 2.9--3.3$\times$ as much energy as VVC at matched PSNR, but 0.60--0.74$\times$ as much at matched bitrates, corresponding to a 26--40\% reduction, with both methods evaluated on the same Apple M3 Pro.
Overall, these results demonstrate that {\name} can achieve low-latency CPU decoding with energy consumption of the same order of magnitude as VVC. }

\rev{Nevertheless, encoding remains time-consuming, making iterative Gaussian optimization an important target for further encoding acceleration.
To accelerate the encoding process, we employ CUDA Graphs to reduce the overhead of repeated GPU kernel launches and CPU--GPU synchronization during iterative Gaussian optimization, which reduces training time by approximately 53\% (2.1$\times$ speedup), while increasing peak GPU memory usage from 748--794 MiB to 1,180--1,258 MiB.
This is because, without CUDA Graphs, buffers are dynamically allocated based on the actual number of Gaussian--tile intersections in each iteration.
However, with CUDA Graphs, we instead preallocate fixed-size buffers based on the maximum intersection count observed during warm-up, avoiding dynamic allocation and the associated CPU--GPU synchronization during replay.
This conservative allocation increases GPU memory usage.
More accurate buffer capacity estimation could further reduce this memory overhead.}

\rev{\textbf{Ablation Study.}
To better understand the contribution of each component, we conduct ablation studies on Gaussian propagation, two-pass encoding, GSP/GSA/DKS, and residual-based quantization, together with additional analyses of Gaussian pruning and scene-change detection.}

 \begin{itemize}
\item \rev{\textit{Gaussian Propagation.}
We first evaluate Gaussian propagation, the core mechanism for exploiting temporal redundancy in {\name}. In the ``anchor-only'' variant, each frame is optimized independently from scratch, whereas ``propagation'' initializes each frame from the reconstructed Gaussian representation of its predecessor. As shown in~\Cref{fig:Ablation_Study_FP}, propagation consistently improves PSNR, MS-SSIM, and VMAF across all Gaussian budgets.}

\begin{figure}[t]
    \centering
    \includegraphics[width=0.25\textwidth]{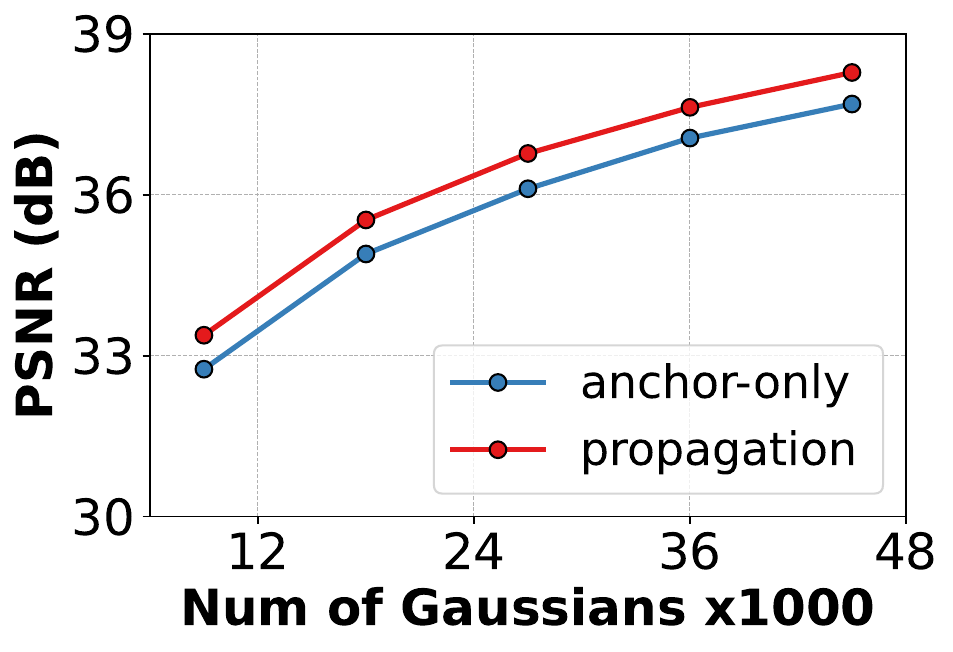}
    \hspace{1mm}
    \includegraphics[width=0.25\textwidth]{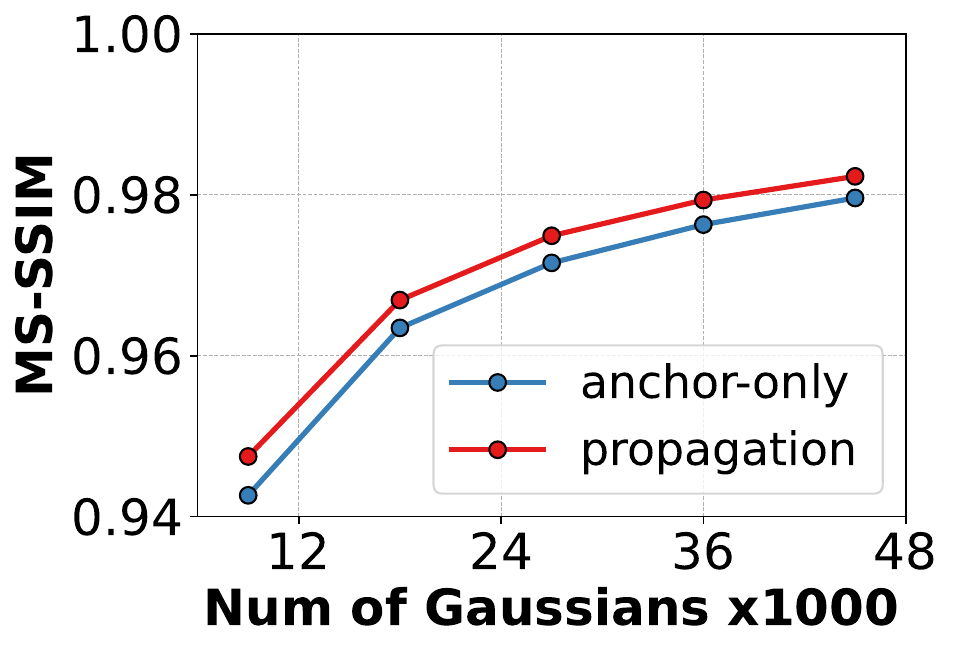}
    \hspace{1mm}
    \includegraphics[width=0.25\textwidth]{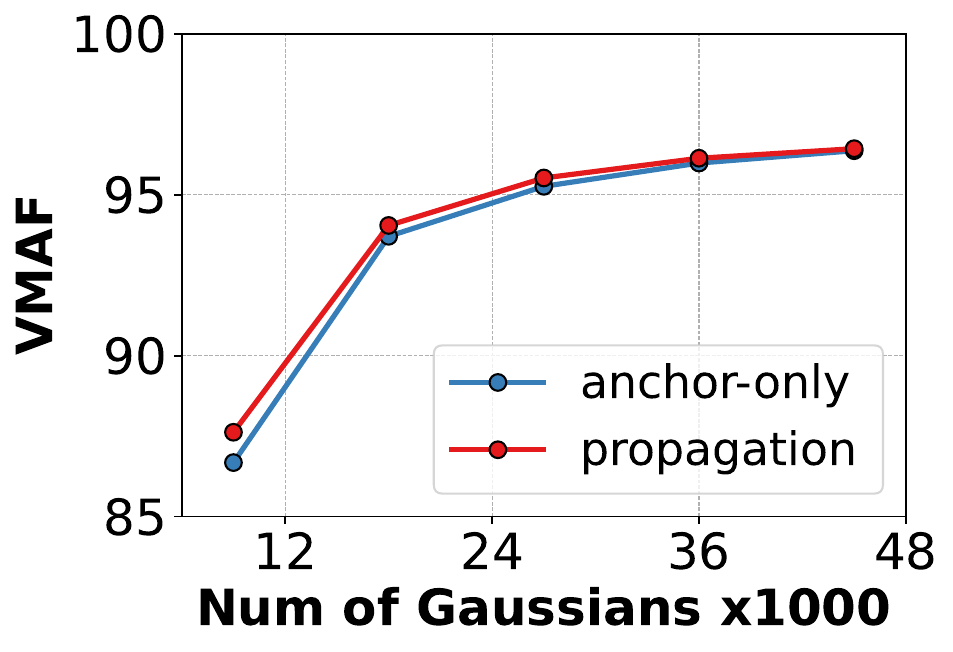}
    \caption{
    \rev{ Effect of Gaussian propagation on PSNR, MS-SSIM, and VMAF across Gaussian budgets before quantization on UVG.}
     }
    \label{fig:Ablation_Study_FP}
\end{figure}

\begin{figure}[t]
    \centering
    \includegraphics[width=0.25\textwidth]{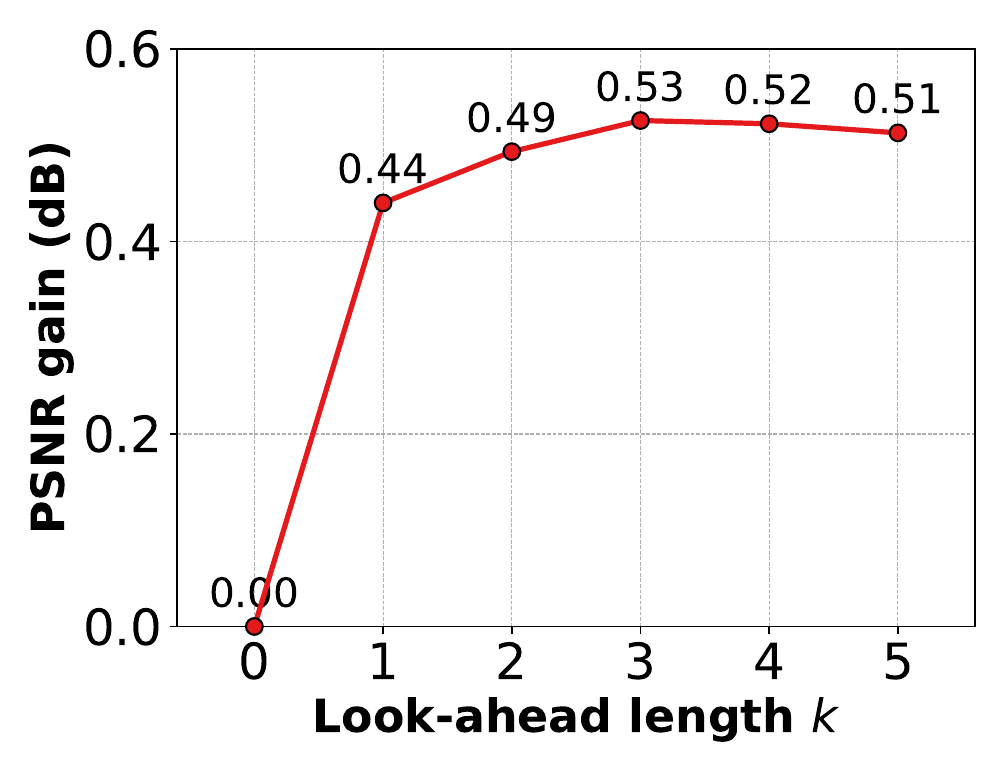}
    \hspace{1mm}
    \includegraphics[width=0.25\textwidth]{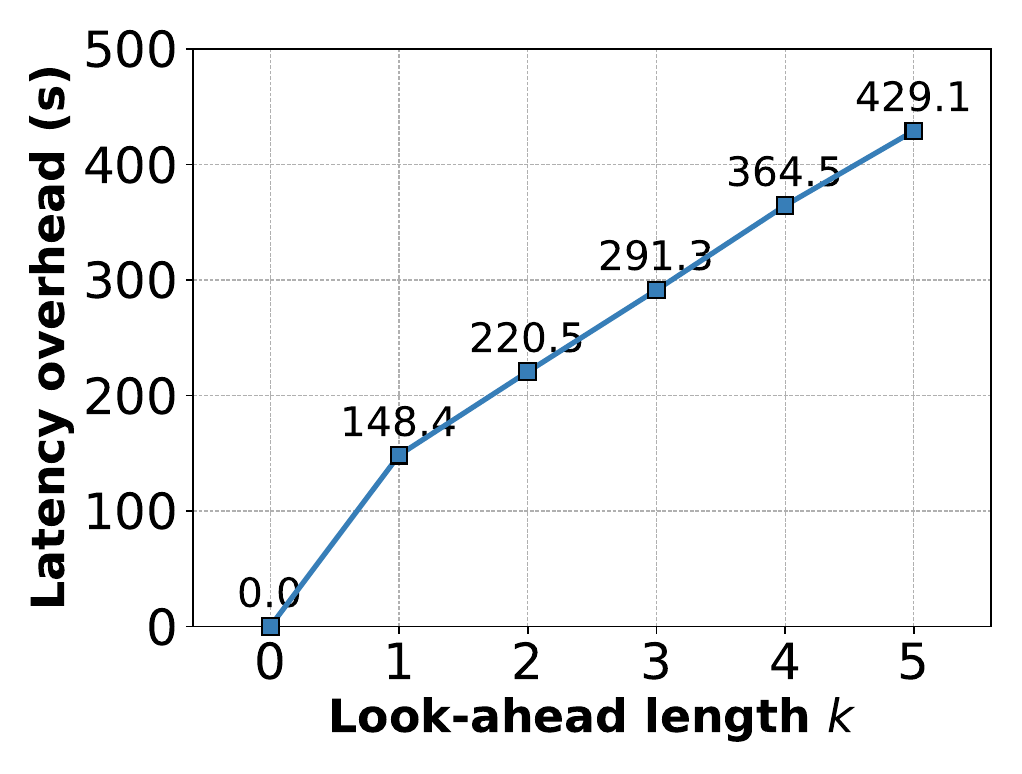}
    \hspace{1mm}
    \includegraphics[width=0.25\textwidth]{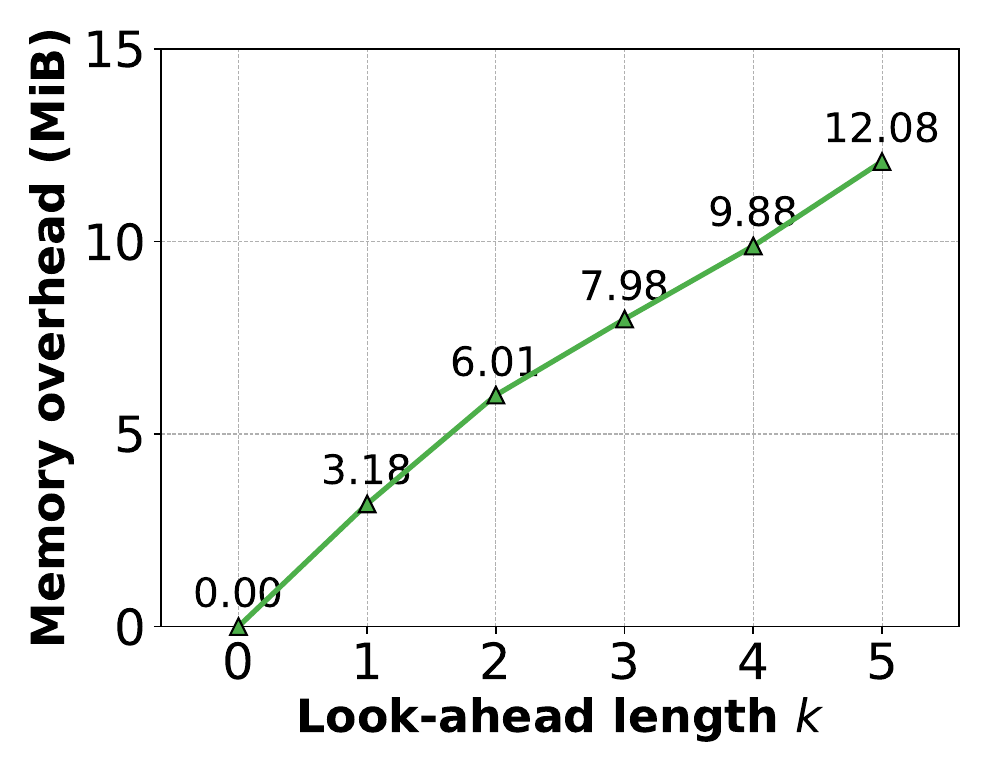}
    \caption{
    \rev{ Ablation study on the look-ahead length of PSNR gain, encoding latency overhead, and peak GPU memory overhead before quantization on UVG.}
     }
    \label{fig:two_pass_encoding_ab_k}
\end{figure}

\rev{\item \textit{Two-pass Training.}
We first study the effect of the look-ahead length $k$ in~\Cref{fig:two_pass_encoding_ab_k}.
Increasing $k$ improves the average PSNR gain from 0.44~dB at $k=1$ to 0.53~dB at $k=3$, after which the gain saturates as more distant frames become less correlated with the anchor.
Meanwhile, compared with $k=1$, the encoding-latency overhead increases by 48.6\% at $k=2$ and 96.3\% at $k=3$, while the additional PSNR gains are only 0.05~dB and 0.09~dB, respectively.
We therefore use $k=1$ by default, as it captures most of the quality gain with lower encoding cost.
In contrast, the memory overhead remains small, increasing from 3.18~MiB at $k=1$ to 12.08~MiB at $k=5$, corresponding to approximately 0.8\% and 3.0\% of the single-pass memory usage, since future-frame representations are optimized consecutively and temporary buffers are reused.
Notably, two-pass encoding introduces no bitstream overhead, as only the final refined anchor representation is quantized and stored.
Then, we evaluate the effect of the two-pass encoding strategy.
On the UVG dataset, we focus on the most challenging sequences, \textit{ReadySetGo} and \textit{YachtRide}, which contain complex textures and plenty of fast motion, providing a stricter setting for evaluation.
In this experiment, we apply two-pass training on every \textcolor{black}{anchor} and compare it against the single-pass variant.
\Cref{fig:Ablation_Study2} reports the performance gap between these two settings.
``w/o second pass'' optimizes anchor frames in a single pass, whereas ``w/ second pass'' applies the proposed two-pass optimization.
As shown in the figure, two-pass training consistently improves both PSNR and MS-SSIM, with more than $0.5$~dB PSNR gain on these sequences.
The improvements on MS-SSIM and VMAF are moderate, because the ``w/o second pass'' baseline already provides near-saturated quality, reaching around 0.99 MS-SSIM and 99 VMAF at 45k Gaussians, which is close to their respective upper bounds (1 for MS-SSIM and 100 for VMAF).}

\begin{figure}[t]
    \centering
    \includegraphics[width=0.25\textwidth]{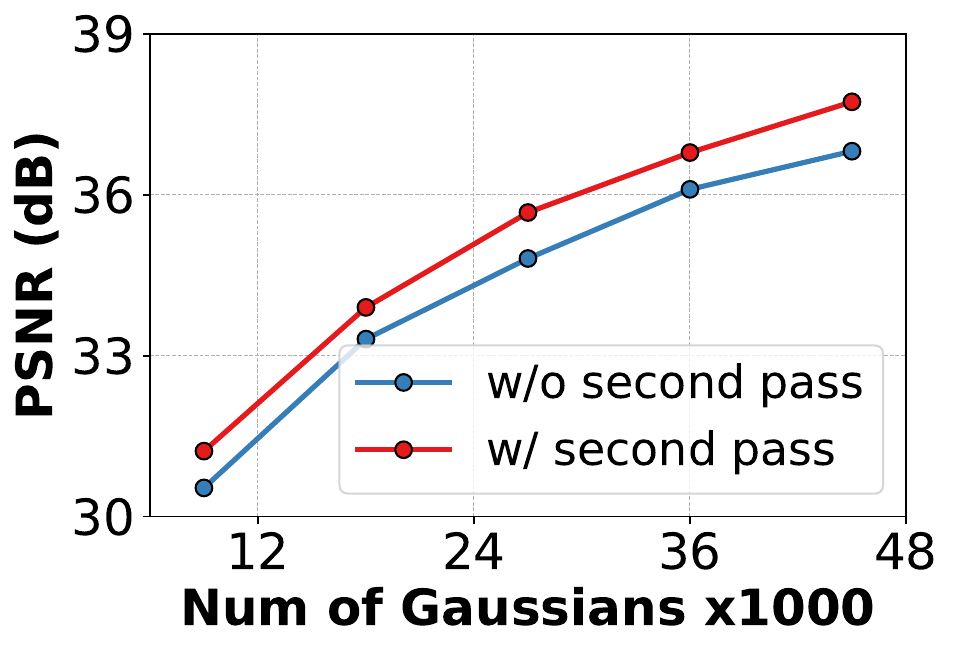}
    \hspace{1mm}
    \includegraphics[width=0.25\textwidth]{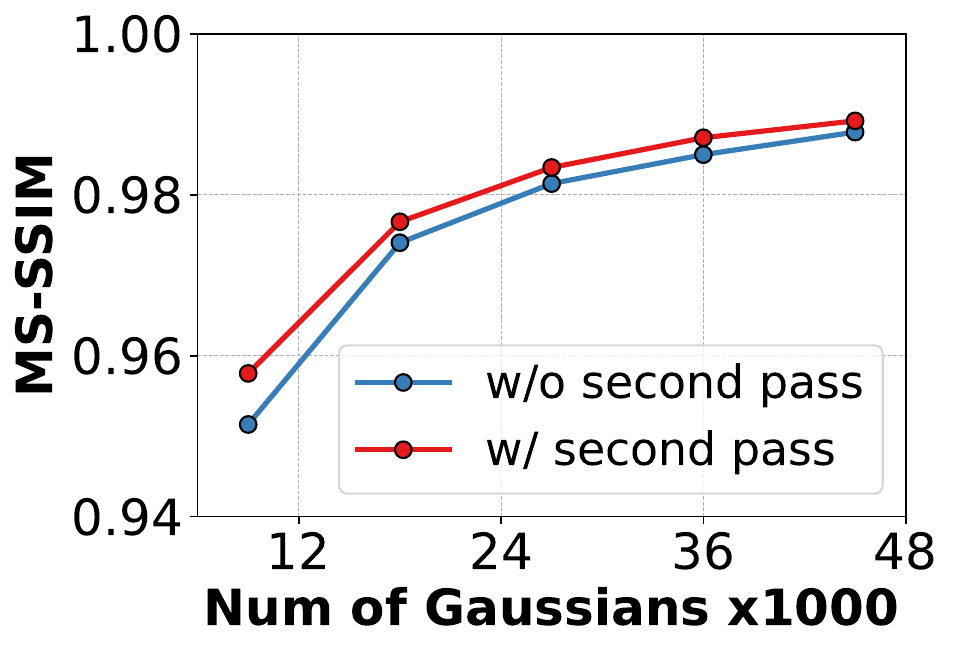}
    \hspace{1mm}
     \includegraphics[width=0.25\textwidth]{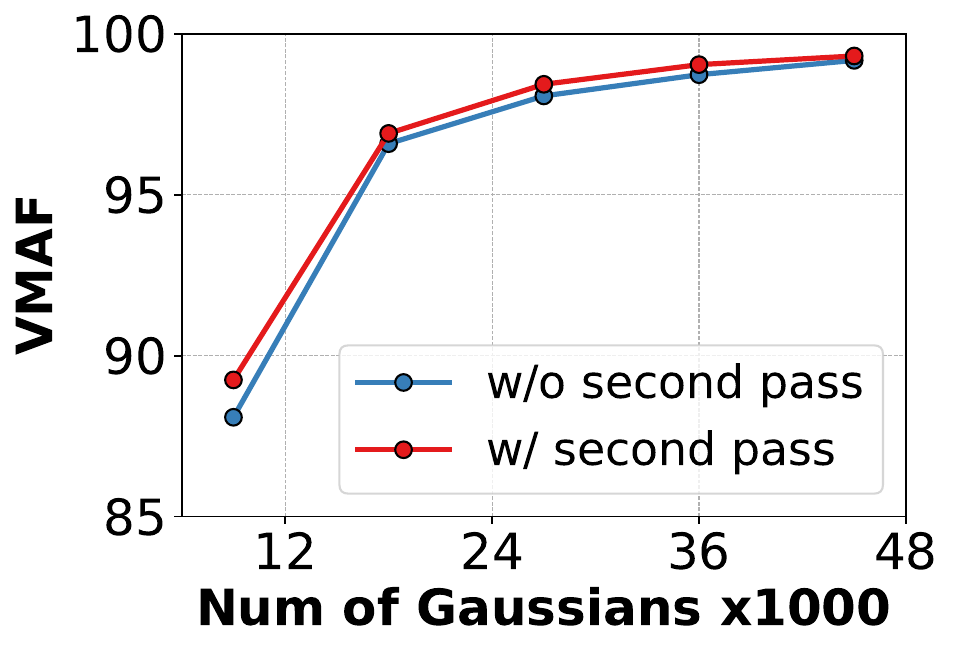}
    \caption{
     \rev{Effect of two-pass encoding on PSNR, MS-SSIM, and VMAF across Gaussian budgets before quantization.}
     }
    \label{fig:Ablation_Study2}
\end{figure}

\item \textit{GSP/GSA/DKS.} To analyze the contribution of Gaussian splat pruning, augmentation, and scene transition detection in {\name}, we compare four variants of our pipeline: (1) \textcolor{black}{propagation frames} only, (2) \textcolor{black}{propagation frames} with GSP, (3) \textcolor{black}{propagation frames} with GSP and GSA, and (4) \textcolor{black}{propagation frames} with GSP, GSA, and DKS. 
\rev{All variants use the same training configuration. As shown in~\Cref{fig:Ablation_Study1}, GSP provides the largest improvement, followed by GSA, while DKS has little effect on MS-SSIM because the evaluated UVG sequences contain no major scene transitions.}

\begin{figure}[t]
    \centering
    \includegraphics[width=0.25\textwidth]{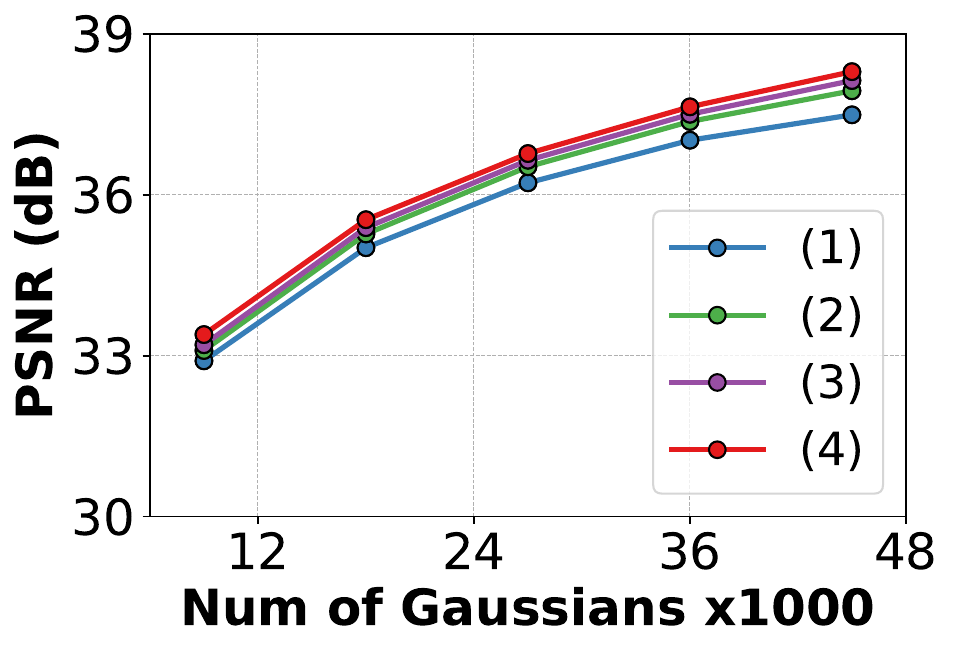}
    \hspace{1mm}
    \includegraphics[width=0.25\textwidth]{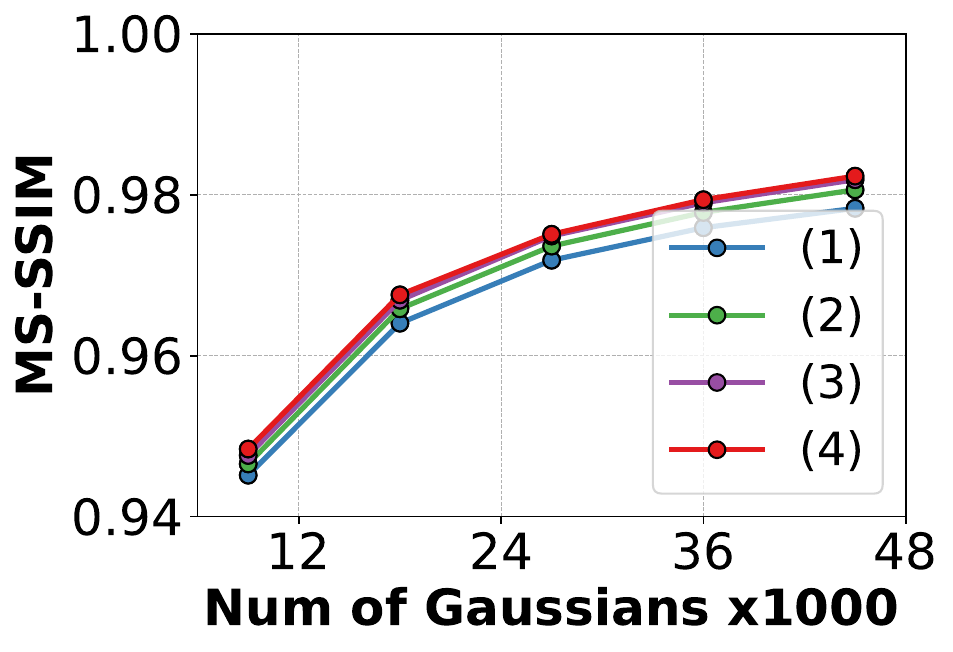}
    \hspace{1mm}
    \includegraphics[width=0.25\textwidth]{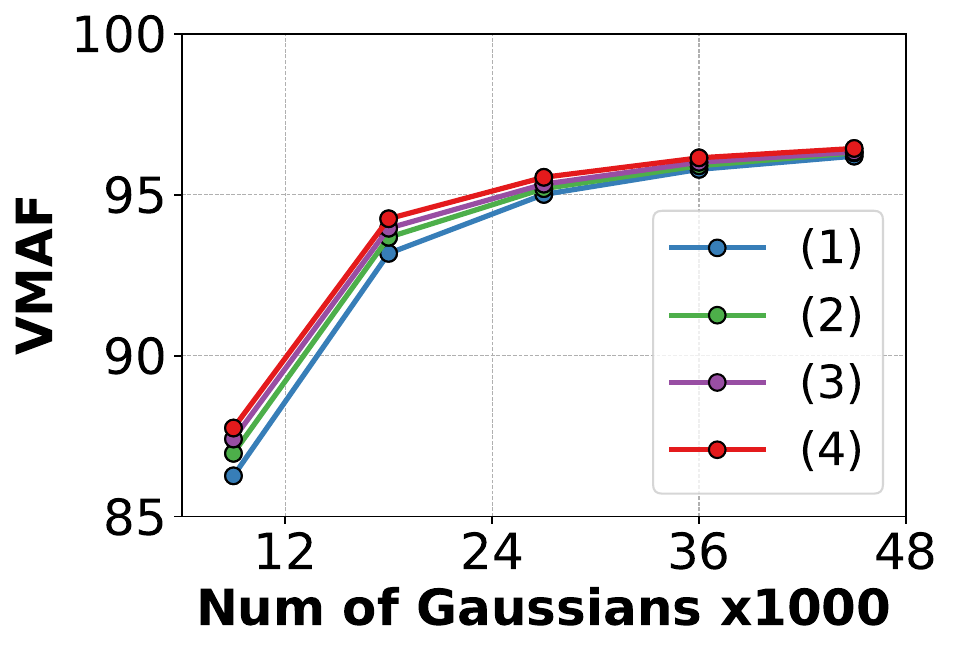}
    \caption{
     \rev{Effects of GSP, GSA, and DKS on PSNR, MS-SSIM, and VMAF across Gaussian budgets before quantization on UVG.}
     }
    \label{fig:Ablation_Study1}
\end{figure}

\rev{\item \textit{Gaussian Pruning.}
We further evaluate the pruning criterion and computational overhead of GSP in~\Cref{fig:Ablation_Study_GSP}.
Compared with post-training random pruning, our GSP, $\mathrm{Contribution}^{\mathrm{FT}}$, improves PSNR by 16.4~dB, showing that GSP avoids indiscriminate pruning that severely damages a converged representation.
Compared with $\mathrm{Random}^{\mathrm{FT}}$, GSP improves PSNR by 0.14~dB while achieving a $1.63\times$ speedup.
This is because removing low-contribution Gaussians causes less disruption to the learned representation and consequently requires fewer optimization steps to recover during fine-tuning.
Post-training $\mathrm{Random}$ pruning is 9.2\% faster than $\mathrm{Contribution}^{\mathrm{FT}}$ because it performs only a one-shot removal after convergence, without contribution evaluation or subsequent fine-tuning.
However, this modest reduction in training time comes at the cost of a substantial reconstruction-quality degradation in both PSNR and MS-SSIM.
Overall, GSP provides a better quality-complexity trade-off by preserving reconstruction quality while substantially reducing the recovery cost associated with random pruning.}

\begin{figure}[t]
    \centering
    \includegraphics[width=0.25\textwidth]{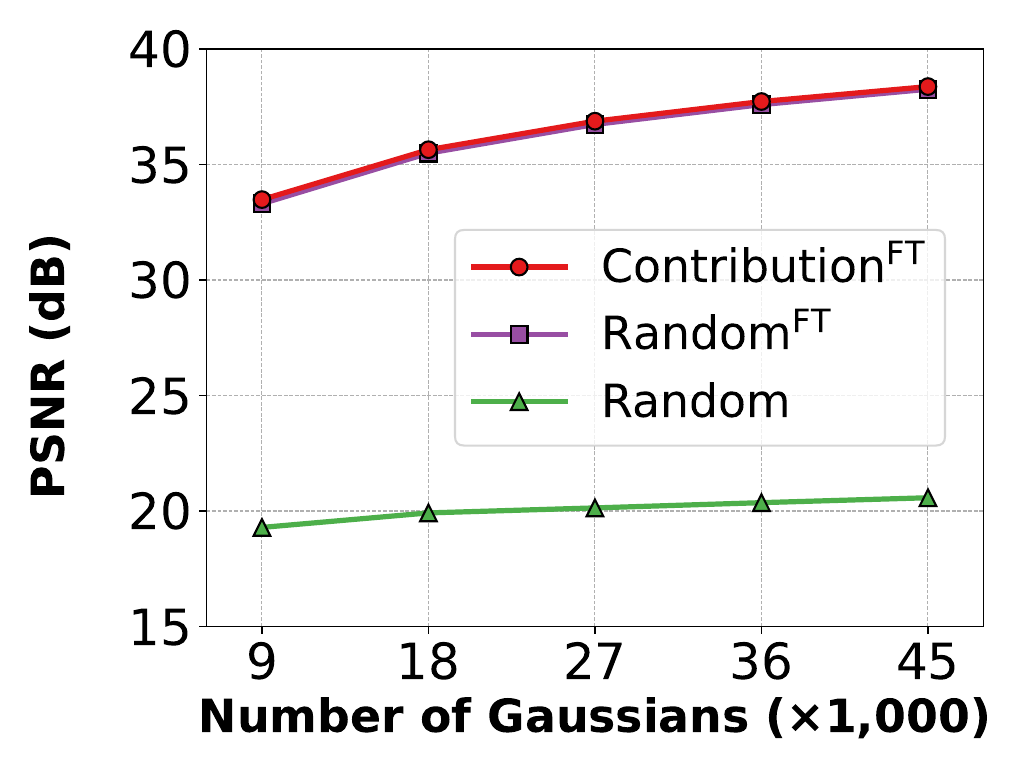}
    \hspace{1mm}
    \includegraphics[width=0.25\textwidth]{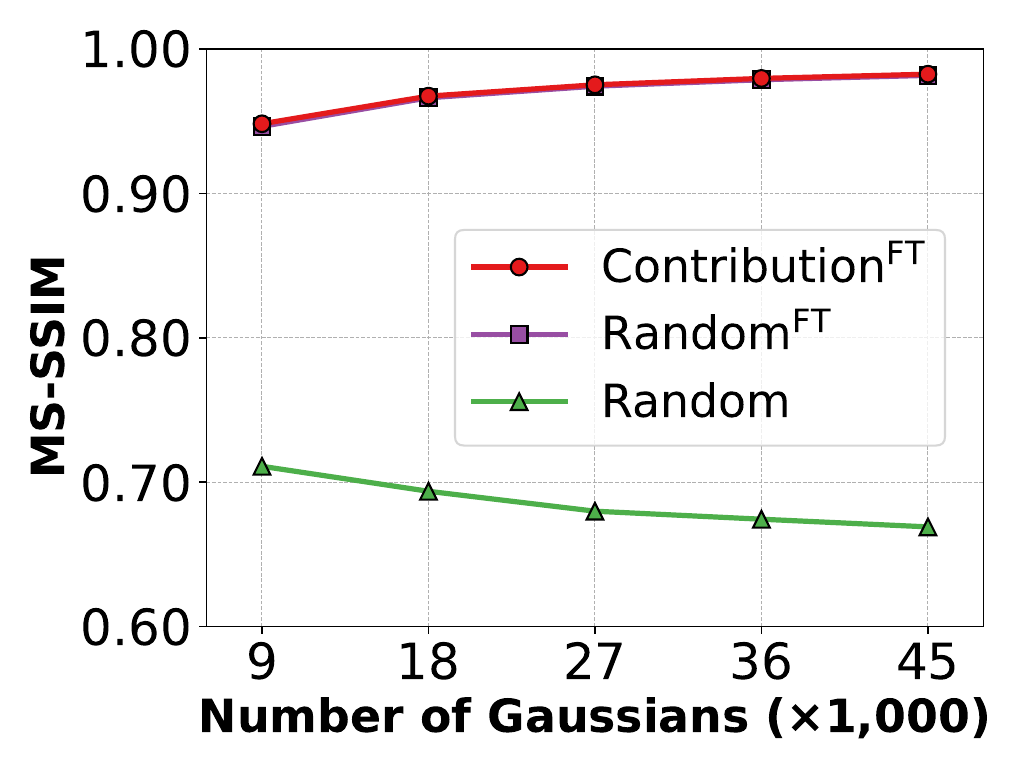}
    \hspace{1mm}
    \includegraphics[width=0.25\textwidth]{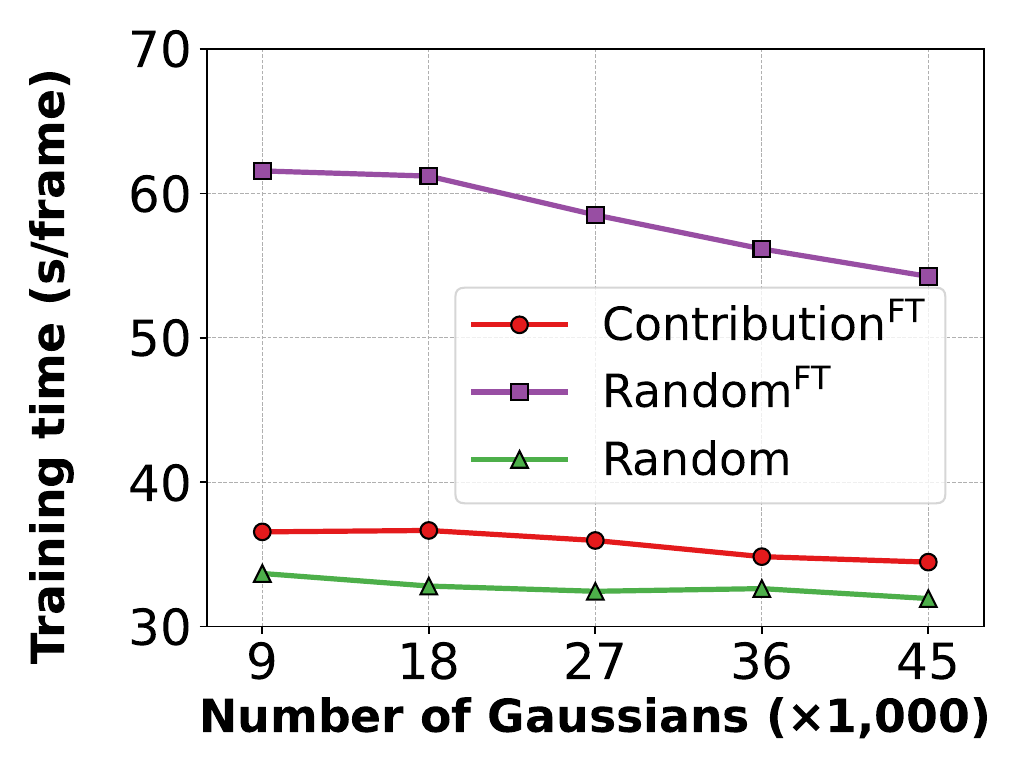}
    \caption{
    \rev{Ablation study of Gaussian pruning strategies on UVG in terms of average PSNR, MS-SSIM, and per-frame training time before quantization. $\mathrm{Contribution}^{\mathrm{FT}}$ denotes GSP (contribution-based pruning with fine-tuning), while $\mathrm{Random}^{\mathrm{FT}}$ and $\mathrm{Random}$ denote random pruning with and without fine-tuning, respectively.}
     }
    \label{fig:Ablation_Study_GSP}
\end{figure}

\item \rev{\textit{Robustness of Scene-Change Detection.}
We evaluate the threshold robustness of DKS alongside two representative frame-domain scene-change detectors: the histogram-based HSV method~\cite{DBLP:conf/iotdcc/SunZ17} and the feature-based ORB method~\cite{DBLP:conf/iccv/RubleeRKB11}.
We concatenate the first 50 frames of the seven UVG sequences into a 350-frame video containing six known scene transitions.
As shown in~\Cref{fig:scene_change_detection}, all three methods achieve 100\% precision and recall over method-specific stable threshold ranges: $[2.25,5.0]$ for DKS, $[0.03,0.598]$ for HSV, and $[0.66,0.98]$ for ORB.
These results show that DKS, like HSV and ORB, can reliably identify hard scene transitions without precise threshold tuning in this setting.
Unlike frame-domain detectors, DKS evaluates whether the current frame is better represented as an anchor or through Gaussian propagation.
Moreover, Gaussians optimized during DKS can be reused for subsequent anchor optimization, avoiding redundant computation at the cost of temporarily retaining Gaussian parameters and optimizer states in GPU memory, without introducing bitstream overhead.
Since our goal is to explore representation-aware anchor selection rather than outperform existing scene-change detectors, we retain HSV and ORB as optional alternatives within our framework.}

\begin{figure}[t]
    \centering
    \includegraphics[width=0.25\textwidth]{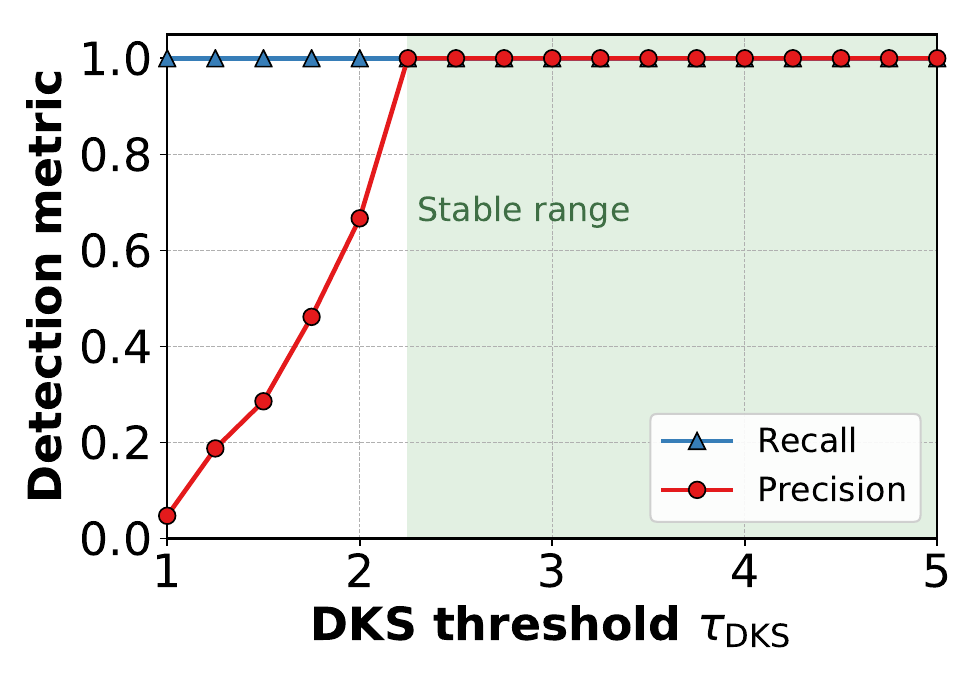}
    \hspace{1mm}
    \includegraphics[width=0.25\textwidth]{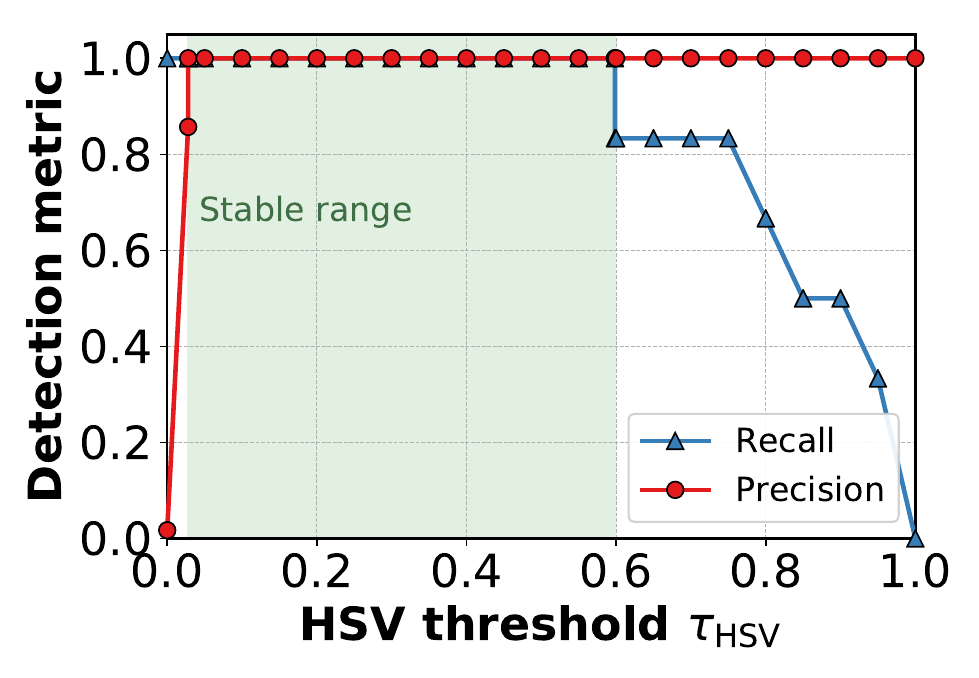}
    \hspace{1mm}
    \includegraphics[width=0.25\textwidth]{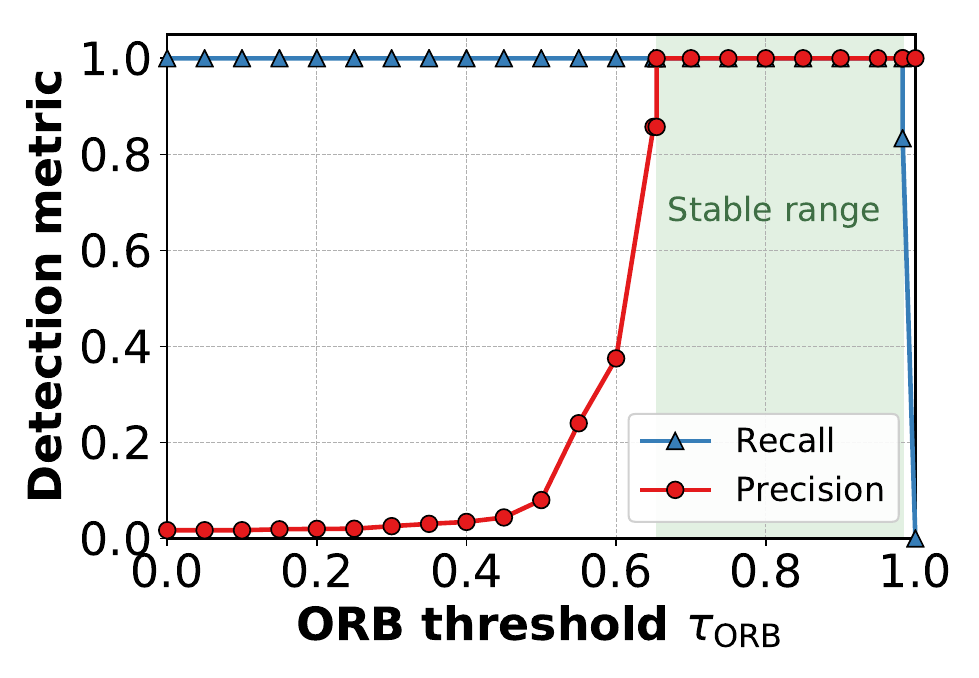}
    \caption{
    \rev{Threshold sensitivity of DKS, HSV, and ORB. We report the precision and recall under varying detection thresholds. The shaded regions indicate stable threshold ranges over which both precision and recall remain 1.}
     }
    \label{fig:scene_change_detection}
\end{figure}

\item \textit{Residual-based Quantization.}
Finally, we conduct an ablation study on the quantization strategy by comparing two variants:
 ``I-only'': all frames are encoded independently, forming an I-frame-only sequence;
``IP'': this corresponds to the residual-based quantization strategy adopted in {\name}, where we track Gaussians’ deltas and decompose each P-frame residual into two components: \rev{intra-coded Gaussian splats and parameter updates of inter-coded splats. As shown in \Cref{fig:Ablation_Study3}, for the ``IP'' variant, temporal changes are represented as low-bit residuals while maintaining almost the same or higher reconstruction quality (PSNR and MS-SSIM), yielding a reduction of more than 34\% in bitstream size compared with the ``I-only'' variant under same PSNR.}
Note that ``IP'' achieves higher VMAF than the ``I-only'' variant with the same Gaussian budget.
This is because VMAF is sensitive to temporal consistency: while ``I-only'' encodes each frame independently and introduces frame-to-frame fluctuations, ``IP'' inherits and refines Gaussian splats across frames through residual encoding, leading to smoother temporal evolution and thus higher perceptual quality.
\end{itemize}

\begin{figure}[t]
    \centering
    \includegraphics[width=0.25\textwidth]{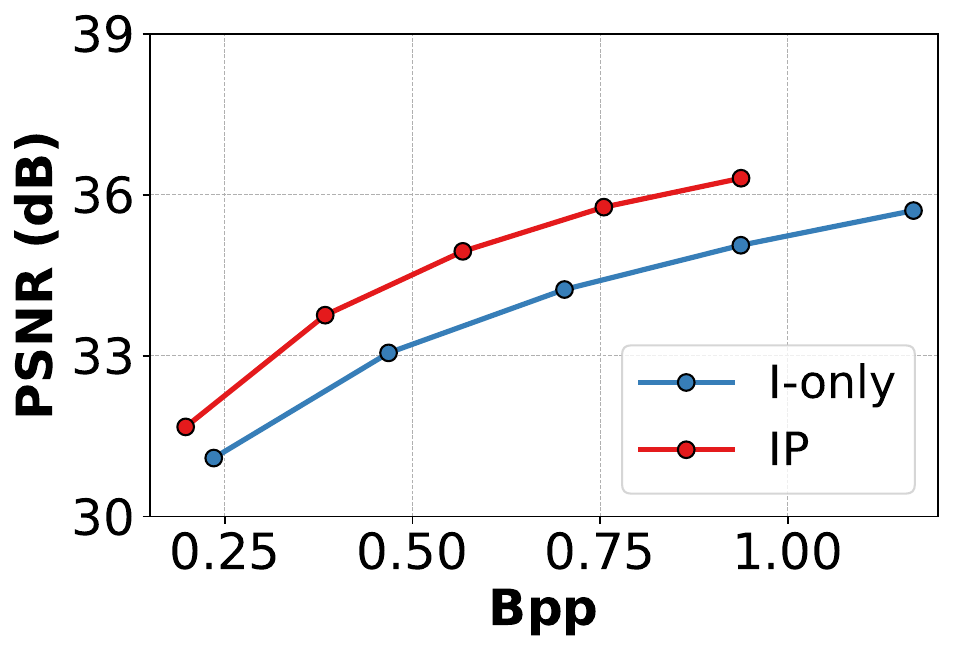}
    \hspace{1mm}
    \includegraphics[width=0.25\textwidth]{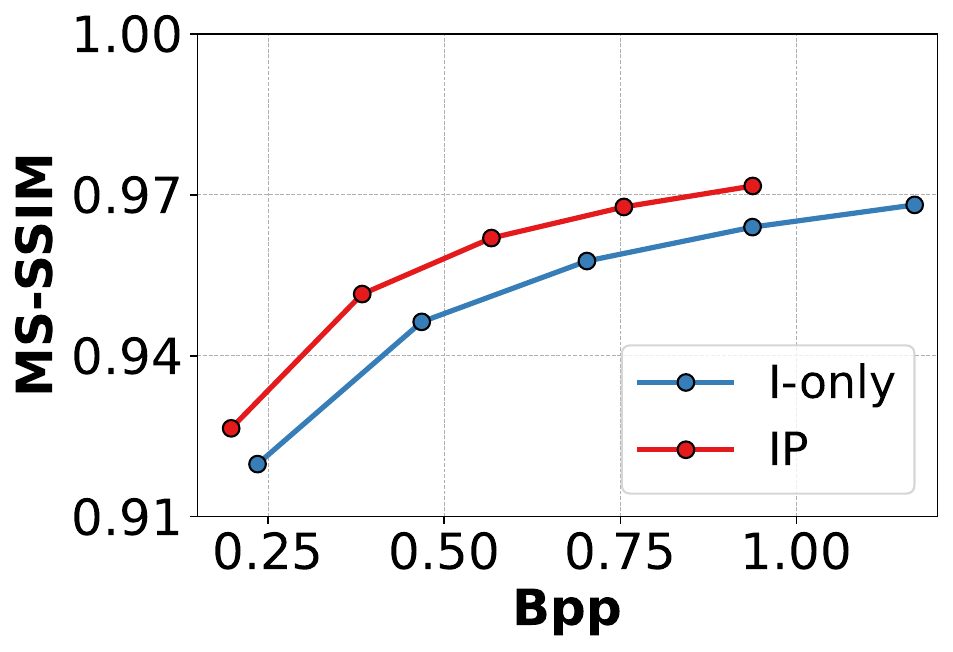}
    \hspace{1mm}
    \includegraphics[width=0.25\textwidth]{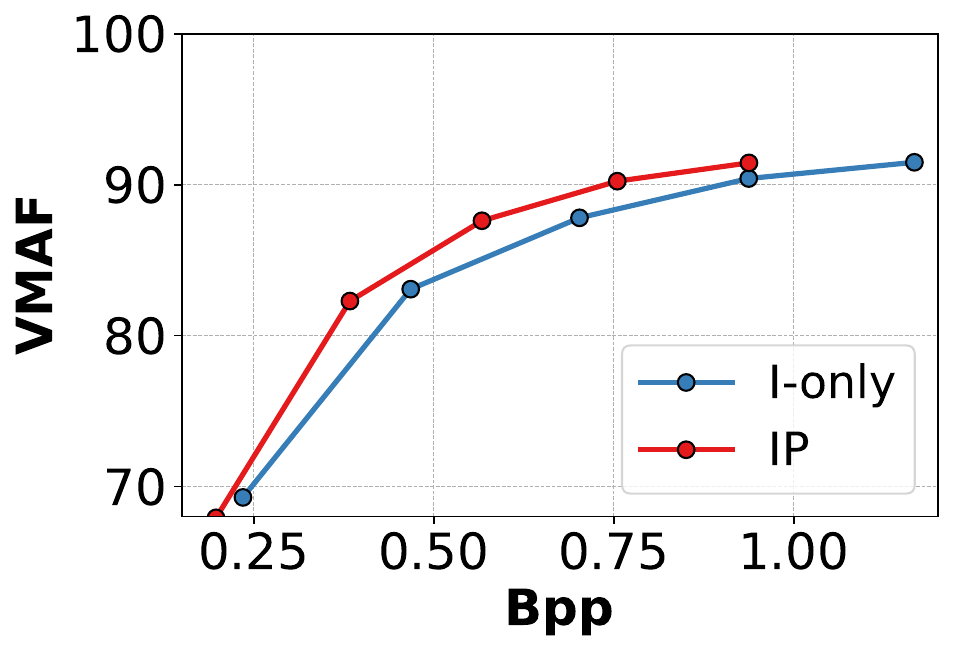}
    \caption{
     \rev{Effect of residual-based quantization on PSNR, MS-SSIM, and VMAF across bitrates on UVG.}
     }
    \label{fig:Ablation_Study3}
\end{figure}

\section{Conclusion}
\rev{{\name} explores 2D Gaussian splats as an alternative representation for video coding.
By optimizing the Gaussian parameters, {\name} achieves rate-distortion performance comparable to widely used implementations of modern video codecs.
Unlike neural-based approaches, its explicit representation provides fine-grained control over compressed size (number of Gaussian splats), reconstruction quality (convergence threshold), and encoding time (number of training iterations).
Furthermore, it can naturally adapt to different resolutions and support fine-grained progressive coding (by adding back the pruned Gaussian splats).}

\rev{A key limitation of {\name} is its high encoding cost, as both Gaussian representation and quantization require extensive iterative optimization.
Future work could reduce the encoding time of {\name} through more efficient optimization and parallel training techniques, benefiting from acceleration techniques in 3DGS~\cite{DBLP:conf/iccv/LiuHND25,DBLP:conf/cvpr/RenWFL26,DBLP:conf/cvpr/ChenJJTLLN25}, as well as feed-forward approaches from 2D Gaussian image reconstruction~\cite{DBLP:journals/corr/abs-2605-20820}, while preserving rate-distortion performance.
Beyond encoding acceleration, {\name} could benefit from more advanced Gaussian compression techniques~\cite{niedermayr2024compressed,navaneet2023compact3D,fan2023lightgaussian,shi2025sketch}, bidirectional prediction, and further hyperparameter tuning.
Building on our previous work on layered progressive Gaussian representations~\cite{shi2025lapisgs}, we have also extended 2D Gaussian video coding to scalable coding in P-GSVC~\cite{DBLP:conf/mmsys/WangSO26}.
Overall, our results provide evidence of the potential of Gaussian splats as an alternative representation for video coding.}

\bibliographystyle{ACM-Reference-Format}
\bibliography{dblpref}
\end{document}